%% file: main.tex
\documentclass[10pt,twocolumn,letterpaper]{article}
\usepackage[pagenumbers]{cvpr}
\usepackage{ulem}
\input{custom_commands}

\definecolor{cvprblue}{rgb}{0.21,0.49,0.74}
\usepackage[pagebackref,breaklinks,colorlinks,citecolor=cvprblue]{hyperref}
\usepackage{floatrow}
\usepackage{bbding}
\usepackage{graphicx}
\usepackage{amsmath}
\usepackage{amssymb}
\usepackage{booktabs}
\usepackage{float}
\usepackage[accsupp]{axessibility} 
\usepackage[title]{appendix}
\usepackage{lipsum}
\usepackage{multirow}

\title{TEDRA: Text-based Editing of Dynamic and Photoreal Actors}

\author{
    Basavaraj Sunagad$^{1}$ \quad
    Heming Zhu$^{1 \ \dag}$ \quad
    Mohit Mendiratta$^{1 \ \dag}$ \quad
    Adam Kortylewski$^{1,3}$ \\
    Christian Theobalt$^{1,2}$ \quad
    Marc Habermann$^{1,2 \ *}$\\
    \\
    $^{1}$Max Planck Institute for Informatics, Saarland Informatics Campus,\\
    $^{2}$Saarbrücken Research Center for Visual Computing, Interaction and AI\\
    $^{3}$University of Freiburg\\
    {\tt\small \{bsunagad, hezhu, mmendira, akortyle, theobalt, mhaberma\}@mpi-inf.mpg.de}
}

\begin{document}
\maketitle
\input{sections/0_abstract}
\footnote[0]{\dag\ Equal contribution.}
\footnote[1]{* Corresponding author.}
\footnote[2]{Project page: \href{https://vcai.mpi-inf.mpg.de/projects/Tedra/}{\color{magenta}{\url{vcai.mpi-inf.mpg.de/projects/Tedra}}}}

\input{sections/1_intro}
\input{sections/2_relatedwork}
\input{sections/3_method}

\input{sections/5_experiments}

\input{sections/7_conclusion}
{
    \small
    \bibliographystyle{ieeenat_fullname}
    \bibliography{main}
}

\clearpage
\setcounter{figure}{0}
\setcounter{table}{0}

\input{sections/X_suppl}

\end{document}

%% file: custom_commands.tex
\usepackage[dvipsnames]{xcolor}
\usepackage{colortbl}

\newcommand{\acronym}{TEDRA}
\newcommand{\trihuman}{TriHuman}

%% file: sections/0_abstract.tex
\begin{abstract}
Over the past years, significant progress has been made in creating photorealistic and drivable 3D avatars solely from videos of real humans. 
However, a core remaining challenge is the fine-grained and user-friendly editing of clothing styles by means of textual descriptions. 
To this end, we present \acronym~ the first method allowing text-based edits of an avatar, which maintains the avatar's high fidelity, space-time coherency, as well as dynamics, and enables skeletal pose and view control.
We begin by training a model to create a controllable and high-fidelity digital replica of the real actor. 
Next, we personalize a pre-trained generative diffusion model by fine-tuning it on various frames of the real character captured from different camera angles, ensuring the digital representation faithfully captures the dynamics and movements of the real person. 
This two-stage process lays the foundation for our approach to dynamic human avatar editing.
Utilizing this personalized diffusion model, we modify the dynamic avatar based on a provided text prompt using our Personalized Normal Aligned Score Distillation Sampling (PNA-SDS) within a model-based guidance framework. 
Additionally, we propose a time step annealing strategy to ensure high-quality edits.
Our results demonstrate a clear improvement over prior work in functionality and visual quality.
\end{abstract}

%% file: sections/1_intro.tex
%
%
\vspace{-1em}
\section{Introduction} \label{sec:introduction}
Digital avatars of real humans play a vital role in various applications, including augmented and virtual reality, gaming, movie production, and synthetic data generation \cite{habermann2020deepcap, li2020volumetric,liu2021neural, peng2021animatable, xu2021hnerf, NNA, zheng2023avatarrex, habermann2023hdhumans, kwon2023deliffas}. However, creating highly realistic and easily animatable avatars presents significant challenges due to the intricate and diverse nature of human geometry and appearance. 
While creating and editing highly realistic avatars is possible, it remains a time-intensive and manual process requiring substantial expertise.
Achieving a specific individual's likeness further adds complexity to this. 

\input{figures/fig_teaser}

In the past years, text-driven image synthesis attracted the attention of the research community, as text is one of the most user-friendly data modalities that can be easily deployed without any expert knowledge.
Thanks to the wide development and adaptation of transformers~\cite{AttentionAllYouNeed,Radford2018ImprovingLU} and diffusion models~\cite{rombach2021highresolution}, several works have shown the ability to edit images in 2D~\cite{brooks2022instructpix2pix,ruiz2022dreambooth} and 3D~\cite{poole2022dreamfusion,instructnerf2023,jain2021dreamfields}, given a text prompt as input. 
While 2D-based methods produce visually convincing results, they most likely can not produce edits that are 3D-consistent. 
In contrast, 3D-based methods~\cite{instructnerf2023,poole2022dreamfusion,jain2021dreamfields} show results on a 3D volume that can be rendered faithfully from an arbitrary camera viewpoint. 
However, such methods are mostly limited to static scenes.

Several methods have been proposed to generate 3D avatars from textual descriptions by distilling the 2D prior of generative models into 3D avatar representations \cite{hong2022avatarclip, huang2023dreamwaltz, kolotouros2023dreamhuman}. 
However, despite the promising results, these methods often fail to adequately capture the dynamic and fine-grained details such as clothing movement.
These aspects are crucial for interactive and dynamic applications.

Recent efforts \cite{10.1145/3618368, shao2023control4d} have focused on modifying dynamic avatars while preserving their underlying motion. However, these approaches are restricted to the upper body \cite{shao2023control4d, 10.1145/3618368} and struggle to generalize to novel poses \cite{10.1145/3618368}.
A significant and open challenge persists in seamlessly applying text-based edits to highly realistic and controllable full-body avatars of real humans. 
The critical requirement is that these edits must maintain spatio-temporal consistency, dynamics, and the high fidelity of the original avatar, all while adhering to user-specified modifications.

In this work, we propose \acronym, the first text-based method for editing the appearance of a dynamic full-body avatar while preserving intricate details (see Fig.~\ref{fig:teaser}).
Our approach assumes a full-body avatar as input, which is trained from multi-view video.
In particular, our work builds upon \trihuman ~\cite{zhu2023trihuman}, 
where the avatar representation is modeled as a signed distance and radiance field anchored to an explicit and deformable mesh template, allowing fast inference of photorealistic human appearance and geometry. 
Through a pre-training stage, a drivable and photoreal digital avatar of the real actor is obtained. 
\par 
Our method enables the text-based editing of such a dynamic volumetric avatar, ensuring spatial and temporal coherence. 
More precisely, we achieve editing through text-based conditional image generation utilizing a diffusion model. 
We contribute several technical advancements to ensure that the editing of \acronym~is authentic, personalized, and maintains visually convincing spatiotemporal consistency. 
Initially, we subsample frames from multi-view videos to capture the avatar's identity and dynamics. We then fine-tune a pre-trained diffusion model with these frames and a unique text identifier to create a personalized generative model that captures the avatar's detailed characteristics. 
Building on this, we introduce Personalized Normal-Aligned Score Distillation Sampling (PNA-SDS), a model-based, classifier-free method inspired by Zhang et al.~\cite{Zhang2022SINESI}. The method employs two latent diffusion models—one personalized and one pre-trained—that perform iterative edits on the dynamic avatar. These diffusion models are conditioned on rendered normals of the avatar to preserve the dynamics while enhancing localized edits. Additionally, noise estimates from both the personalized and pre-trained diffusion models are strategically combined at specific timesteps to optimally balance the original avatar characteristics with the intended modifications.

Furthermore, to prevent over-saturation artifacts, we introduce a windowed annealing strategy, which gradually reduces the influence of the personalized diffusion model and enables high-frequency edits. 
In summary, our contributions are:
\begin{itemize}
    \item We introduce \acronym, a method for editing dynamic 3D full-body avatars based on textual input. Our approach combines neural volumetric scene representations with text-driven diffusion models. This allows for precisely editing dynamic digital avatars while preserving detailed wrinkle patterns and ensuring seamless animatability.
    \item We propose a novel technique, termed as Personalized Normal Aligned Score Distillation Sampling (PNA-SDS), facilitating high-quality personalized editing while maintaining the integrity of dynamics.
    \item We present windowed time-step annealing for score distillation from text-to-image diffusion models, preventing over-saturation artifacts and achieving high-quality edits.
\end{itemize}
We conduct a comprehensive evaluation of our method, employing, both, subjective and numerical assessments through a user study and comparisons with related techniques. 
The results demonstrate that our approach not only generates a wide range of text-based edits but also maintains the integrity of the initial identity. 
Additionally, we showcase animations of the edited avatars, further highlighting the temporal coherency and superior performance of our method compared to other relevant approaches.

%% file: figures/fig_teaser.tex
%
%
\begin{figure*}[t]
  \centering
  \includegraphics[width=0.99\linewidth]{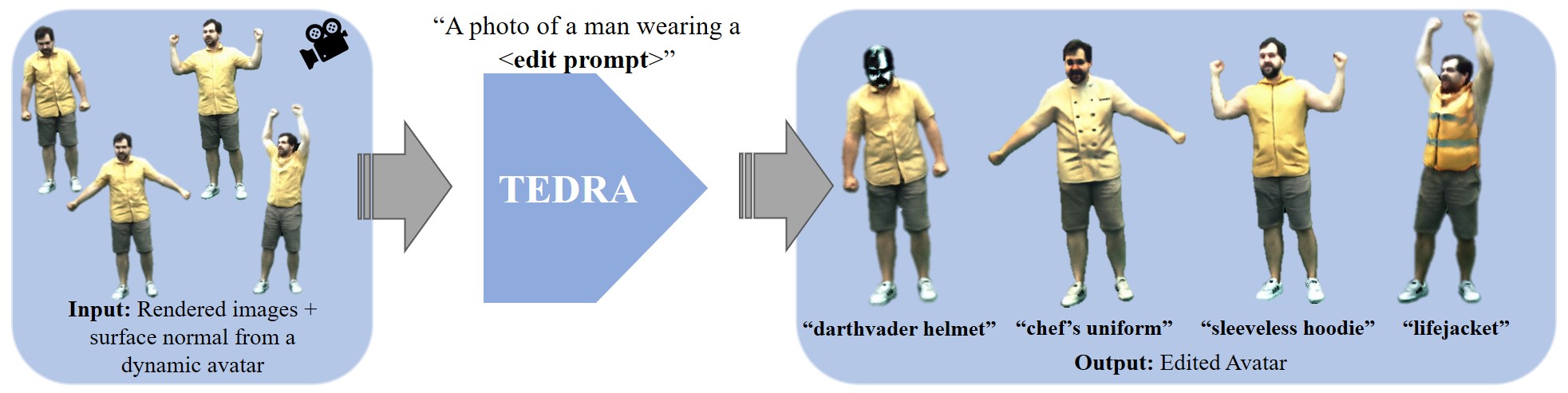}
  \caption
  {
  We propose a method for text-based editing of dynamic and photoreal actors (TEDRA).
  Our approach edits a pre-trained neural 3D human avatar according to a user-defined text prompt.
  Importantly, we preserve the original dynamics and view consistency of the digital avatar while also satisfying the desired edit. 
  }
  \label{fig:teaser}
\end{figure*}
%
%

%% file: sections/2_relatedwork.tex
%
%
\section{Related Work} \label{sec:related}
%
\input{figures/fig_overview}
%
%
\noindent\textbf{Diffusion-based Text-to-3D Generation and Editing.}
In the realm of computer graphics, the synthesis of 3D assets from textual descriptions has emerged as a captivating area of research. 
These assets are generated using Score Distillation Sampling (SDS), a technique introduced in DreamFusion~\cite{poole2022dreamfusion}, which enables the generation of 3D content from textual inputs by lifting text-to-image diffusion models to 3D domain.
Fantasia3d~\cite{Chen_2023_ICCV} proposes a text-to-3D content creation method by disentangling the modeling and learning process of geometry and appearance. 
Meanwhile, Hi-FA~\cite{zhu2023hifa} introduces a novel timestep annealing approach that progressively reduces the sampled timestep throughout a single-stage optimization process. 
InstructNerf2Nerf~\cite{instructnerf2023} employs an image-conditioned diffusion model for editing NeRF scenes with text instructions. 
Given a Neural Radiance Field (NeRF)~\cite{mildenhall2021nerf} representation of a scene and respective multi-view images, the method uses an image-conditioned diffusion model~\cite{brooks2023instructpix2pix} to iteratively edit the input images while optimizing the underlying scene. 
In contrast, Re-PaintNerf~\cite{zhou2023repaintnerf} starts with the semantic selection of the object to be modified, followed by guiding the NeRF model using a pre-trained diffusion model. 
All these works have in common that they create \textit{static} scenes, which neither support direct animation nor modeling of piece-wise rigid articulated objects and respective dynamics, e.g. motion-induced clothing deformations. 
%
%
\par
\noindent\textbf{Diffusion-based Text-to-3D Avatar Generation.}
In the field of text-driven 3D avatar generation, a series of methodologies have been proposed recently, each tackling a specific challenge. For instance, AvatarVerse~\cite{zhang2023avatarverse} and AvatarCraft~\cite{jiang2023avatarcraft} focus on generating avatars based solely on textual input. In contrast, DreamAvatar~\cite{cao2023dreamavatar} prioritizes creating avatars with controllable poses and body types. HumanNorm~\cite{humannorm2023} takes a unique approach, enhancing the perceived realism of 3D avatars by focusing on how 2D information translates to 3D geometry.  Several methods, including DreamWaltz~\cite{huang2023dreamwaltz}, DreamHuman~\cite{kolotouros2023dreamhuman}, and TADA~\cite{liao2023tada}, combine text-driven generation with pre-built body models to create animatable avatars. Additionally, ZeroAvatar~\cite{weng2023zeroavatar} and TeCH~\cite{huang2023tech} aim to improve the overall quality and detail of generated avatars, while HaveFun~\cite{havefun} tackles the problem of reconstructing avatars from just a few photos. 
However, all the above works rely on a parametric body model to create a 3D avatar. 
When dealing with diverse avatars that deviate significantly from the parametric model, employing the original skinning weights in such instances will result in animations perceived as unrealistic. 
Secondly, and most importantly, they do not model the dynamics of the surface and appearance, i.e. wrinkles and cast shadows.

Recent works such as DynVideo-E~\cite{liu2023dynvideoeharnessingdynamicnerf} and Control4D~\cite{shao2023control4d} focus on 4D editing, ensuring temporal and spatial consistency by utilizing an underlying radiance field representation. However, these methods do not prioritize preserving the details and fidelity of the underlying avatar, nor do they adequately capture deformations such as wrinkles. 
A closely related work, AvatarStudio~\cite{10.1145/3618368} introduces View-Time SDS for maintaining consistency in editing the underlying facial avatar across multiple views. Nevertheless, this method encounters challenges in generalizing to facial motions that it has not encountered before and faces limitations in real-time rendering. Additionally, the use of view-specific prompts in AvatarStudio leads to an entanglement of camera and identity views, restricting its applicability to datasets where the person's movement is limited.
%
%

\par
\noindent\textbf{Drivable 3D Avatars.}
Modeling high-fidelity, dynamic 3D-clothed human avatars has been an emerging topic in recent years. 
Here, we focus on the works related to the modeling of drivable 3D avatars, which take \textit{only skeletal poses} and \textit{virtual camera views} as input at inference time as they are most closely related.

Previous research on drivable avatars can be divided into two streams: mesh-based and hybrid approaches. 
Mesh-based methods~\cite{Volino2014,casas14,habermann2021,shysheya2019textured} represent the shape and appearance of dynamic characters with drivable template meshes with static (dynamic) textures.
However, the rendering quality is bounded by the resolution of the underlying mesh template.

To improve the quality of both the generated geometry and rendering, hybrid approaches articulate implicit fields~\cite{henzler2019escaping,sitzmann2019deepvoxels,sitzmann2019srns} or radiance fields~\cite{mildenhall2020nerf,zhang2020nerf++} with the explicit shape proxies, i.e., 3D skeletons, parametric human body models~\cite{loper15,STAR:2020,SMPL-X:2019,TotalCapture2018}, or person-specific template meshes ~\cite{liu2020neural,kwon2023deliffas,zhu2023trihuman,habermann2023hdhumans}.
A prevalent research trend~\cite{su2021nerf,weng2020vid2actor,chen2021animatable, 2021narf,ARAH,bergman2022generative,jiang2022instantavatar,li2022tava,su2022danbo,Feng2022scarf,SHERF} focuses on modeling dynamic humans by mapping the posed space to a pose-agnostic canonical space.
To better model the pose-dependent appearance of humans, recent studies ~\cite{liu2021neural, peng2021animatable, xu2021hnerf, NNA, zheng2023avatarrex, habermann2023hdhumans, kwon2023deliffas} incorporate motion-aware residual deformations in the canonicalized space.
Among them, Neural Actor~\cite{liu2021neural} and HDHumans~\cite{habermann2023hdhumans} leverage the texture space of the human body mesh as local features to model dynamic human appearances. Nevertheless, both methods require approximately 5 seconds to render a single frame.
\trihuman ~\cite{zhu2023trihuman} achieves real-time rendering and geometry generation through a deformable tri-plane anchored on the motion-controllable template mesh.
The rendering and geometry quality is on par with, or even better than, the previous offline methods and significantly excels the real-time methods. 
We, therefore, take it as our underlying drivable avatar representation.
Nevertheless, all these approaches exhibit a limitation as they do not allow for text edits and solely animate the clothing presented in the video.

%% file: figures/fig_overview.tex
%
%
\begin{figure*}[t]
\centering  \includegraphics[width=0.98\textwidth]{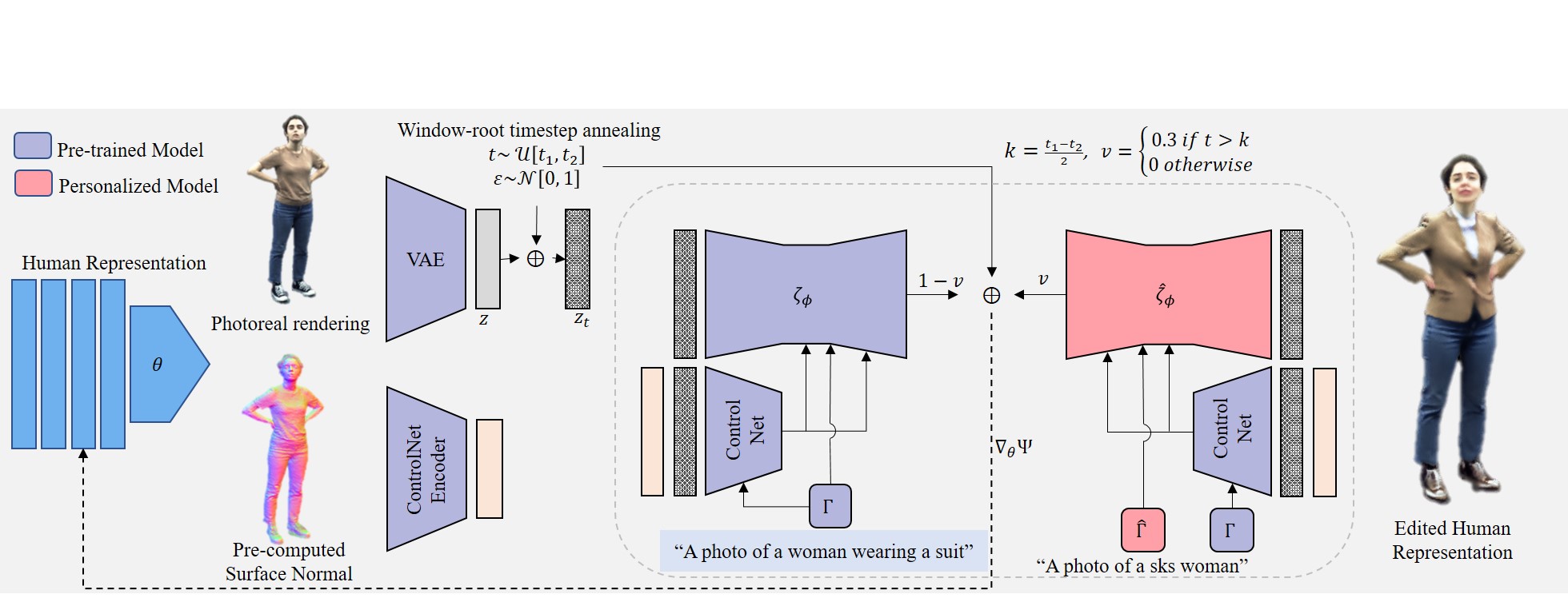}
  \caption
  {
An overview of our approach for text-driven editing of dynamic and photoreal avatars (TEDRA). 
Our approach starts with a pre-trained \trihuman~model as the base human representation.
Then, we leverage a fine-tuned diffusion model in conjunction with our proposed Personalized Normal Aligned Score Distillation Sampling (PNA-SDS). 
The PNA-SDS technique then computes a normal aligned model-based score distillation sampling loss to optimize the human representation towards the edit prompt while preserving the subject's characteristics. 
This process is further enhanced by incorporating an annealing mechanism, which gradually refines the editing process.
}
\label{fig:pipeline}
\end{figure*}
%
%

%% file: sections/3_method.tex
%
%
\section{Method} \label{sec:method}
We aim to edit a pre-trained 3D human avatar, learned from multi-view video data, using textual prompts that specify desired changes, such as "Man wearing a hoodie" (see Fig.~\ref{fig:pipeline}). 
The main challenge is to maintain the avatar's overall characteristics, transfer the original clothing dynamics, and achieve the specified edits.
\par 
We leverage the recent state-of-the-art neural 3D avatar representation, \trihuman ~\cite{zhu2023trihuman}, due to its high geometric and visual quality as well as its real-time performance. 
The challenge lies in preserving intricate dynamics, such as wrinkles and other details, from the pre-trained human avatar during the editing process. 
To achieve this, we propose a novel score distillation-based approach, which effectively leverages the edit capabilities of large text-guided latent diffusion models (LDM)~\cite{StableDiff,rombach2021highresolution} to coherently modify the motion-aware geometry and appearance generated by \trihuman. 
Next, before we introduce TEDRA (Sec.~\ref{sec:tedra}), we first discuss the necessary foundations, i.e., the avatar representation, LDMs, and Score Distillation Sampling (Sec.~\ref{sec:prelim}). 
%
%
\subsection{Preliminaries} \label{sec:prelim}
%
%
\par 
\noindent{\textbf{Avatar Representation.}}
Among the existing methods, we choose \trihuman~\cite{zhu2023trihuman} as the human representation for its ability to generate high-quality, motion-aware, and coherent geometry and appearance of dynamic humans in real-time.
\par 
As a hybrid representation, \trihuman~ firstly adopts an explicit, skeleton-driven human template mesh to depict the human character's coarse geometry.
To generate the template mesh, it learns the non-rigid deformation of the human template mesh in the canonical space in a graph-to-graph translation manner~\cite{habermann2021}. 
The non-rigid deformed canonical template meshes are then transformed to the posed space via Dual Quaternion Skinning~\cite{kavan2007skinning}.
\par 
While the template mesh captures the motion-aware dynamics of the human characters, the fidelity of the appearance and geometry is bounded by the resolution of the template mesh. 
To this end, \trihuman~ introduced a deformable volume, parameterized with a Tri-plane, anchored on the deformable character's texture space.
Given a spatial sample $\mathbf{x}_j$ along marching rays in the posed space, \trihuman~ non-rigidly maps the sample $\mathbf{x}_j$ to the texture volume bridged by the pose-deformed template mesh through inverse skinning.
The respective position of $\mathbf{x}_j$ in the texture volume is denoted as $\mathbf{u}_j$.
The sampled Tri-plane's features, denoted as $\mathbf{F}_{j,f}$ at position $\mathbf{u}_j$ are further fed into two shallow MLPs, i.e., the shape MLP $\mathcal{H}_\mathrm{sdf}$, and the color MLP $\mathcal{H}_\mathrm{col}$, to produce the corresponding SDF $s_{j,f}$ and color value $\mathbf{c}_{j,f}$.
%
\begin{align}
\mathcal{H}_\mathrm{sdf}(\mathbf{F}_{j,f}, p(\mathbf{u}_j) = s_{j,f}, \mathbf{q}_{j,f}\label{eq:our_final_sdf} \\
\mathcal{H}_\mathrm{col}(\mathbf{q}_{j,f}, \mathbf{n}_{j,f}, p(\mathbf{d})) = \mathbf{c}_{j,f}.\label{eq:our_final_col}
\end{align}
%
where $\mathbf{q}_{j,f}$ denotes the motion-aware local shape features, $\mathbf{n}_{j,f}$ is the surface normal, $p$ indicates positional encoding~\cite{mildenhall2020nerf}, $f$ denotes the frame index, and $\mathbf{d}$ is the ray direction. 
Lastly, unbiased volume rendering~\cite{wang2021neus} is adopted to integrate the ray samples and generate the final rendering. 
We refer to the original work~\cite{zhu2023trihuman} and our supplemental document for more details.
%
%
\par 
\noindent{\textbf{Latent Diffusion Models.}}
In Latent Diffusion Models (LDMs) \cite{rombach2021highresolution}, an encoder function $\mathcal{E}$ maps a high-dimensional data point $\mathbf{x}$, e.g., an image, into a lower-dimensional latent representation $\mathbf{z}=\mathcal{E}(\mathbf{x})$.
This latent space representation is then subject to a diffusion process \cite{ho2020denoising, nichol2021improved}, a Markov chain of $T$ steps. Each step adds a small amount of Gaussian noise, gradually transforming the data into a noise distribution. Mathematically, this process can be described as:
%
\begin{equation}\label{eq:noising_func}
\mathbf{z}_t = \sqrt{\alpha_{t}} \mathbf{z} + \sqrt{1 - \alpha_{t}} \boldsymbol{\epsilon}, \boldsymbol{\epsilon} \sim \mathcal{N}(0, I). 
\end{equation}
%
where the diffusion timestep $t$ ranges from \(1 \leq t \leq T\). The reverse diffusion process in LDMs aims to gradually denoise the latent representation to generate new data samples. The reverse process is modeled by a neural network $\boldsymbol{\epsilon}_\phi$, where $\phi$ represents the trainable parameters of the network. This network learns to predict the noise $\boldsymbol{\epsilon}$ added at each step, enabling the model to reverse the diffusion process. Diffusion models, akin to various other generative models, are inherently capable of capturing conditional distributions denoted as $p(\mathbf{x} \vert \mathbf{c})$, where $\textbf{c}$ represents a conditioning variable. The learning process for the conditional latent diffusion models then minimizes
%
\begin{equation}
\mathbb{E}_{\boldsymbol{\epsilon} \sim \mathcal{N}(0,1), t}\left[\|\boldsymbol{\epsilon} - \boldsymbol{\epsilon}_\phi (\mathbf{z}_t, t, \mathbf{c}) \|_{2}^{2}\right].
\end{equation}
%
%
\par 
\noindent{\textbf{Score Distillation Sampling (SDS).}}
Score Distillation Sampling \cite{poole2022dreamfusion} optimizes an implicit 3D representation from textual descriptions to generate a view-consistent scene using pre-trained 2D text-to-image diffusion models. The approach constructs the scene through a differentiable image parameterization, employing a differentiable generator $\mathcal{G}$ that produces 2D images $\textbf{x}$ from 3D scene parameters $\boldsymbol{\theta}$.
The method utilizes a combination of a pre-trained and personalized diffusion model to derive a score function ${\Psi_{\boldsymbol{\phi}}}(\textbf{x}_t, \mathbf{y}, t)$. This score function estimates the noise $\boldsymbol{\epsilon}$, given the noisy image $\textbf{x}_t$, text embedding $\textbf{y}$, and noise level $t$. The score function is crucial for determining the gradient direction for the scene parameter updates $\boldsymbol{\theta}$. The gradient, essential for updating these parameters, is computed as
%
\begin{equation}
\nabla_{\boldsymbol{\theta}} \mathcal{L}_{\text{SDS}}(\phi, \mathbf{x}) = \mathbb{E}_{t, \epsilon} \left[ \mu(t)({\Psi_{\boldsymbol{\phi}}}(\mathbf{x}_t; \mathbf{y}, t) - \boldsymbol{\epsilon}) \mathbf{\frac{\partial x}{\partial \boldsymbol{\theta}}} \right].
\end{equation}
%
Here, $\mu(t)$ is a weighting based on the diffusion timestep $t$.
%
%
\subsection{Proposed Method} \label{sec:tedra}
We start by employing the pre-trained \trihuman~model to generate images displaying diverse views and poses. 
Subsequently, we fine-tune the pre-trained Stable Diffusion model~\cite{Rombach2021HighResolutionIS} on these generated images, utilizing an identity-specific prompt (Sec.~\ref{sec:fine_tune}). 
To ensure the accurate editing of the pre-trained avatar, we introduce our novel PNA-SDS loss (Sec.~\ref{sec:naip}). 
This loss is computed through model-based classifier-free guidance~\cite{ho2022classifierfree}, guided by the normal-aligned ControlNet~\cite{zhang2023adding}. 
Finally, we introduce our window-root timestep annealing strategy (Sec.~\ref{sec:annealing}), specifically designed for diffusion-guided text-to-3d editing. 
%
%
\subsubsection{Fine-tuning the Latent Diffusion Model} \label{sec:fine_tune}
In dynamic full-body editing, the main challenge is preserving the body's original characteristics, such as identity, cloth deformation details, and motions, rather than completely altering its appearance. 
DreamBooth~\cite{ruiz2022dreambooth} addresses identity-specific image generation by fine-tuning the Latent Diffusion Model (LDM) with an identity-specific prompt and a few images (4-5) of the subject. 
While effective for generating novel images, it is not suitable for consistently editing articulated and animatable avatars. 
Unlike AvatarStudio's~\cite{10.1145/3618368} view-time fine-tuning scheme for short-sequence facial edits with 8-10 images, editing animatable full-body avatars is more complex and requires the LDM to manage novel poses and extensive view-dependent appearances, necessitating long sequences with a large number of frames for training. 
Furthermore, the fine-tuning strategy struggles with large datasets to generate accurate view-time token samples.
\par 
To address this, we propose a more comprehensive fine-tuning strategy that encompasses all conceivable poses and viewpoints, aiming to accurately represent the full spectrum of human surface dynamics and appearances. 
We render multi-view and multi-pose images denoted as $[\textbf{x}_i; i \in \{1,...,n\}]$ from the pre-trained \trihuman~ model for fine-tuning the UNet and the text encoder of the LDM. 
The fine-tuning process is further refined by incorporating an identity-specific token alongside a class noun within the prompt. 
This results in a structured prompt of the form 'a photo of a sks man/woman', where 'sks' is the identity-specific token.
\par 
We then fine-tune the UNet, denoted as $\hat{\zeta}_{\phi}$ and text encoder $\Gamma$, so the LDM can regenerate an image $\textbf{x}_i$ from an initial noise map $ \boldsymbol{\epsilon} \sim \mathcal{N}(0, I)$ and a conditioning vector $\textbf{s} = \Gamma(\textbf{P})$, derived using a text encoder, conditioned on a text prompt $\textbf{P}$. 
The fine-tuning of $\zeta_{\theta}$ and $\Gamma$ is supervised with a squared error loss function and a class-specific prior preservation loss~\cite{ruiz2022dreambooth}. 
The squared error loss for denoising a variably-noised image or latent code, given by $\textbf{z}_{t,i} = \alpha_t \mathcal{E}(\textbf{x}_i) + \beta_t\boldsymbol{\epsilon}$, is expressed as 
\begin{equation}
    \mathbb{E}_{x_i, s_i, \zeta, t} \left[ w_t \|\hat{{\zeta}}_{\phi}(\textbf{z}_{t,i}, \textbf{s}) - \mathcal{E}(\textbf{x}_i) \|^2
_2 \right],
\end{equation}
%
where $\alpha_t, \beta_t,$ and $w_t$ control the noise schedule. 
\par 
After fine-tuning the LDM, we use a pre-trained normal-aligned ControlNet~\cite{zhang2023adding} encoder conditioned with a null prompt to control the virtual camera view and skeletal pose in the generated images. 
%
%
\subsubsection{Personalized Normal-aligned Model-based Score Distillation (PNA-SDS)} \label{sec:naip}
To enable subject-consistent, view, and pose-dependent edits, we compute the edit score with model-based classifier-free guidance~\cite{ho2022classifierfree} for updating the pretrained \trihuman{} model. 
This is done by interpolating the noise estimates from the fine-tuned model $\hat{\zeta}_\phi$ and pre-trained stable diffusion model $\zeta_\phi$, respectively, at specific timesteps. 
Our 3D human representation can generate surface normal maps, which guide the diffusion models via our normal-aligned pre-trained ControlNet. 
We adopt a parameter $k$ based on the diffusion timestep $t$ to determine whether or not to interpolate between the scores, which effectively balances between edits and identity. 
\par 
Given a rendered image $\textbf{x}$, we obtain its latent representation $\textbf{z}$ using the VAE of the stable diffusion: $\textbf{z} = \mathcal{E}(\textbf{x})$. 
We then uniformly sample a diffusion timestep $t \sim \mathcal{U}(t_1, t_2)$, where $t_1$ and $t_2$ represent the upper and lower limits of the diffusion noise timestep $t$. 
These limits are introduced in the next paragraph, where we discuss the annealing process. 
We apply this sampled timestep to introduce noise to the input latent, resulting in a noised latent $\textbf{z}_t$.
\par 
Let $\textbf{c}$ represent the embedding of the editing text prompt (e.g., 'a photo of a man wearing a hoodie'), $\hat{\textbf{c}}$ represent the text embedding of the prompt for the identity (e.g., 'a photo of a sks man'), and $\textbf{n}$ (Eq.~\ref{eq:our_final_col}) represents the normal map rendered with \trihuman.
Then, we can obtain the editing score as follows:

\begin{equation}
    \footnotesize
    \begin{aligned}
        \boldsymbol{\Psi}(\mathbf{z}_t, t, \mathbf{c}, \hat{\mathbf{c}}, \mathbf{n}) 
        & = w((1-v) \boldsymbol{\zeta}_\phi(\mathbf{z}_t, \mathbf{c}, \mathbf{n}) + v\hat{\boldsymbol{\zeta}}_\phi(\mathbf{z}_t, \hat{\mathbf{c}}, \mathbf{n})) \\
        &+(1-w) \boldsymbol{\zeta}_\phi\left(\mathbf{z}_t, \mathbf{n}\right),
    v = \begin{cases} 
    0.3 & \text{if } t > k, \\
    0 & \text{otherwise.}
    \end{cases} \label{equ:pna_sds}
    \end{aligned}
\end{equation}
%
%
where $w$ is the overall guidance weight and $v$ stands for the model guidance weight. 
We use Eq.~\ref{equ:pna_sds} for the noise estimate and $k$ represents a threshold timestep for using a personalized diffusion model. 
It is important to note that incorporating surface normals as a condition for both text-guided and null prompts is essential to maintain spatial relationships and surface orientations within the estimated score. 
Our Personalized Normal Aligned-SDS, combined with a comprehensive fine-tuning strategy, excels in generalizing edits to novel views and poses not seen in the fine-tuning dataset. 
In Fig.~\ref{fig:ablation}, we show the effectiveness of Personalized Normal Aligned-SDS compared with other SDS variants.
%
%
\subsubsection{Windowed Root Timestep Annealing} \label{sec:annealing}
Similar to previous studies~\cite{Chen_2023_ICCV, zhu2023hifa, wang2023prolificdreamer}, our experimental results indicate that Score Distillation Sampling (SDS) encounters notable challenges related to over-saturation and loss of fine details, particularly when a large timestep \( t \) is randomly selected. 
In diffusion models, the higher timesteps correspond to the semantics of the image, while the lower timesteps correspond to finer details~\cite{patashnik2023localizing}. 
Thus, for an editing task, it is crucial to establish the edit semantics early on and then move to add finer details.

Several annealing strategies have been proposed~\cite{Chen_2023_ICCV, zhu2023hifa, wang2023prolificdreamer}, but none of them are suitable for an editing task. 
HiFA~\cite{zhu2023hifa} proposed a square root annealing strategy for selecting the diffusion timestep \( t \) based on the iteration step, directly correlating the diffusion process' progression with the training iteration. 
However, random sampling of \( t \) is crucial for model-based classifier-free guidance as per Eq.~\ref{equ:pna_sds}. 
When \( t \) is deterministically chosen, as in HiFA, we face the following limitations:
\begin{itemize}
    \item With a fixed threshold \( k = 600 \), model-based guidance ceases after \( t > k \), leading to a loss of identity due to the lack of influence from the personalized model in later iterations (refer to PNA-SDS+HiFA annealing in Sec.~\ref{sec:ablations}).
    \item Constant model-based guidance (\( k = 0 \)) restricts edit flexibility and results in blurry artifacts.
\end{itemize}
Thus, it is important to stochastically sample \( t \) to balance identity preservation and edit flexibility.

To address this, we introduce a \textit{windowed square root annealing strategy} specifically designed to modulate the annealing of timesteps while allowing random sampling. 
This approach ensures a more controlled and gradual progression of timesteps during the training process.

Given the total number of iterations \( N \) and the current iteration \( \tau \), along with a specified window size \( w \), our annealing values are:
\begin{equation}
    t_1 = t_{\text{max}} - (t_{\text{max}} - t_{\text{min}}) \times \sqrt{\frac{\tau}{N}}, \,
    k = t_{\text{1}} - \frac{w}{2}, \,
    t_2 = t_1 - w.
    \label{time_step_annealing}
\end{equation}
Here, \( t_{\text{max}} \) and \( t_{\text{min}} \) represent the maximum and minimum diffusion timesteps, respectively. 
The window \([t_1, t_2]\) is the range within which \( t \) is randomly sampled. The parameter \( k \) signifies the timestep for using the personalized diffusion model. 
This dynamic window adapts throughout the annealing process, facilitating a balance between establishing edit semantics at higher timesteps and adding fine details at lower timesteps, thereby mitigating issues of oversaturation.

Moreover, the weight \( v \) of the personalized diffusion model is annealed to enhance the faithfulness of the edits to the input prompt. 
This approach not only improves the semantic integrity during initial higher timesteps but also ensures the preservation of fine details as the process progresses to lower timesteps.

%% file: sections/5_experiments.tex
%
%
\section{Experiments} \label{sec:results}
Our evaluation focuses on three key aspects: 
1) Ensuring alignment with the target text prompt while maintaining the subject's inherent characteristics; 
2) The capability to produce 3D consistent edits for high-quality free-view rendering; 
3) The temporal coherency of the generated edits, enabling dynamic replay and skeleton animations.
\paragraph{Dataset.} \label{sec:dataset}
We conduct experiments on one subject (wearing shorts and a shirt) of the DynaCap dataset~\cite{habermann2021}, which is recorded in a calibrated multi-camera studio.
In addition, we captured three more subjects in various clothing in a similar studio setup. 
For training \trihuman, we obtain skeletal motion using markerless motion capture~\cite{captury}, and foreground masks using background matting~\cite{BMSengupta20}.
For more details, we refer to the supplemental document.
%
%
\begin{figure*}[t]
    \centering    \includegraphics[width=0.98\textwidth]{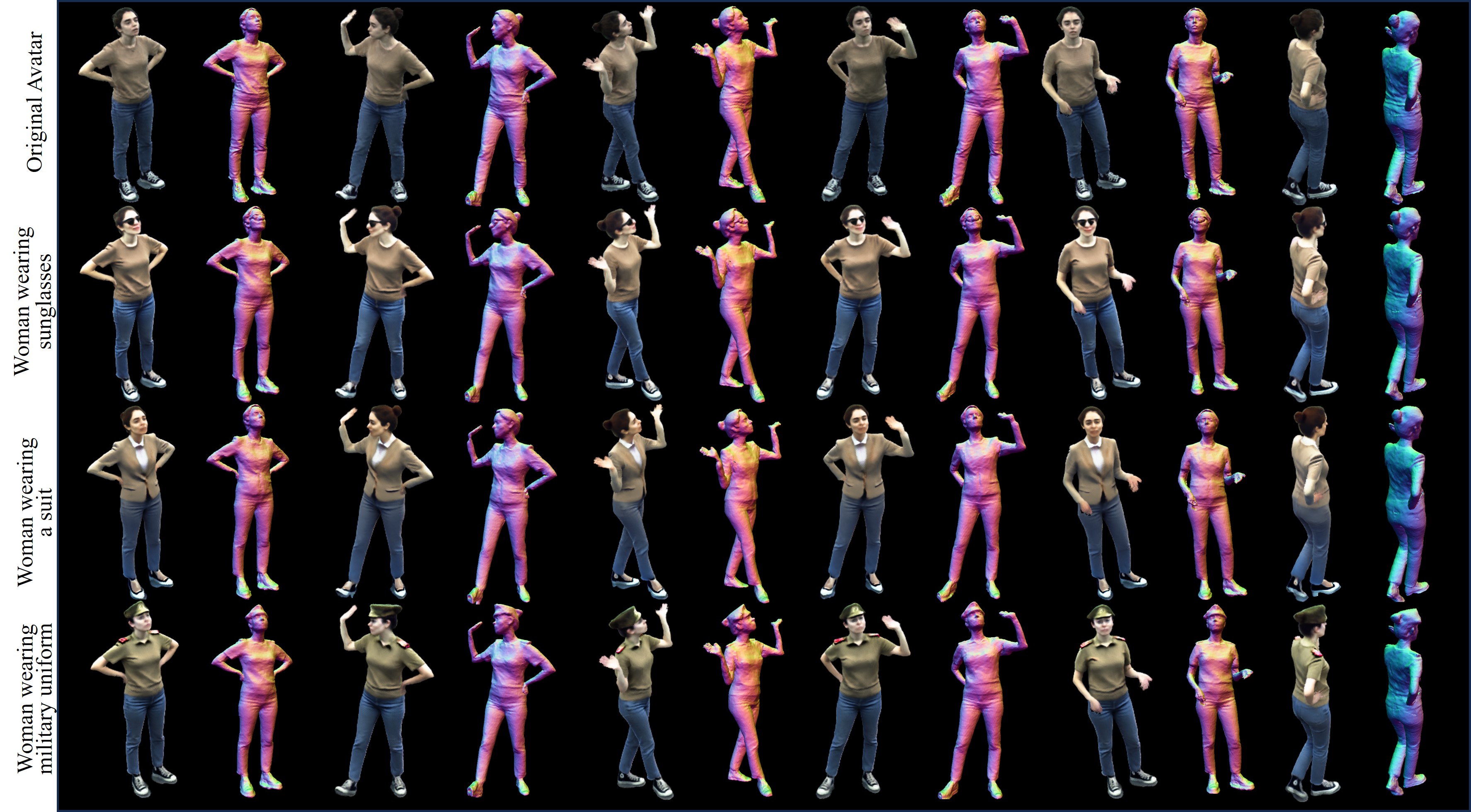}
    \vspace{-7pt}
    \caption{\textbf{Qualitative Results.} We present the text-based visual editing results and the underlying geometry. Our method generates compelling text-driven visual edits, ensuring 3D and temporal consistency while altering appearance and geometry. 
    We recommend the readers to \textbf{zoom in} for better viewing of the details.
    }
    \label{fig:results_id1}
\end{figure*}
%
\input{figures/fig_comparisons}

\subsection{Qualitative Results} \label{sec:quali_results}
Fig.~\ref{fig:results_id1} presents the text-based edits upon a single identity with various poses generated by TEDRA.
The results yield by TEDRA shine in the following aspects: 
\textbf{(1) Localized and precise:} As illustrated in Fig.~\ref{fig:results_id1} ("Women wearing sunglasses"), TEDRA yields fine-grained edits on the target region while preserving the clothing wrinkle details from the original avatar.
\textbf{(2) Versatile and expressive:} As shown in Fig.~\ref{fig:results_id1} ("Women wearing a suit"), TEDRA can modify not only the appearance but also the style of the outfits.
The edited outfits significantly differ from the originals in, both, style and appearance.
\textbf{(3) 3D and temporal coherent:} TEDRA edits, both, the appearance and the underlying geometry of the clothed human avatar to align with the given text prompt, as evident in the representations of Fig.~\ref{fig:results_id1} ("Women wearing military uniform"). 
The edits seamlessly integrate with the drivable avatar and remain coherent across various poses.
Please see the supplemental materials for more results on novel poses and views.

\subsection{Quantitative Evaluation} \label{sec:user_study}

We compare our method with two recent 3D text-based editing approaches: InstructNeRF2NeRF~\cite{instructnerf2023} and AvatarStudio~\cite{10.1145/3618368}. To ensure a fair comparison, we use the same training data for InstructNeRF2NeRF and apply AvatarStudio’s full-head editing strategy to our full-body avatar. Fig.~\ref{fig:comparison} shows the outcomes under different text prompts.

InstructNeRF2NeRF exhibits poor alignment with target prompts and may alter the original identity by editing incorrect regions. AvatarStudio produces blurry results that lack fine details like wrinkles and clothing dynamics. In contrast, our method retains fine details and dynamics, producing more coherent edits aligned with the text.

Given the difficulty of quantitative comparison against ground truth in this setting, we adopted a user study approach from AvatarStudio~\cite{10.1145/3618368}. 
Each session featured four side-by-side videos: the original input and randomly shuffled outputs from three text-driven methods. 
We evaluated three identities with three different prompts, totaling nine videos, each 10-15 seconds long. 
Participants were asked to respond to the following questions:

\begin{itemize}
    \item Q1: Which method better preserves the identity of the input sequence (subject consistency)?
    \item Q2: Which method better adheres to the provided textual prompt (prompt preservation)?
    \item Q3: Which method better maintains the animations and dynamics of the original motion (temporal consistency)?
    \item Q4: Which method performs better overall, considering the three aspects above?
\end{itemize}

Tab.~\ref{tab:userstudy} presents the user study results from 25 participants, with our method consistently receiving the highest ratings. 
For overall quality (Q4), our method was preferred 78.4\% of the time, compared to AvatarStudio’s 10.4\% and InstructNeRF2NeRF’s 11.2\%. We also evaluate CLIP Text-Image Direction Similarity and FID scores. Tab.~\ref{tab:quantitative} shows that our method outperforms prior works in these metrics by a large margin.

\begin{figure}[t]
   \centering
   \includegraphics[width=0.99\linewidth]{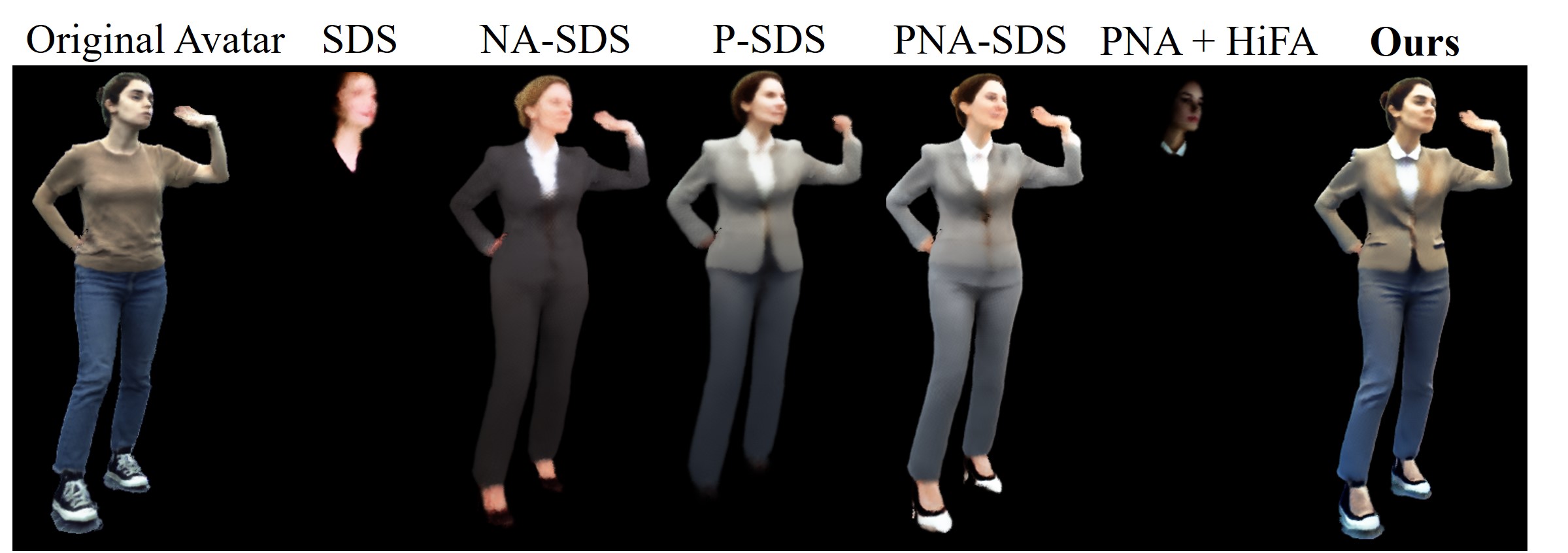}
    \vspace{-7pt}
   \caption{\textbf{Ablation Study.} Our full method achieves visually plausible edits and preserves the dynamics of the original character.}
   \label{fig:ablation}
\end{figure}

\input{tables/quantitative}


\subsection{Ablations} \label{sec:ablations}
In Fig.~\ref{fig:ablation} and Tab.~\ref{tab:clip_ablation}, we conduct ablative studies to assess the effectiveness of our design choices.

Initially, we establish the necessity of incorporating normals as a condition for, both, the edit-prompt and the null-prompt, we use a pre-trained diffusion model \cite{Rombach2021HighResolutionIS} and ControlNet ~\cite{zhang2023adding} (Eq. ~\ref{equ:pna_sds} with $v=0$), which we term as \textbf{NA-SDS}. 
As shown in Fig.~\ref{fig:ablation}, aligning the score with normals crucially helps to preserve geometric details that are lost with standard \textbf{SDS}.

Next, we highlight the importance of adopting the personalized model to preserve the identity. 
Here, we compute the score using our proposed loss, but we exclude the normals to highlight the impact of the personalized model (Eq.~\ref{equ:pna_sds} with $n=\oslash$) termed as \textbf{P-SDS}.
We observe an improvement in the appearance and overall structure of the edits compared to vanilla \textbf{SDS}. 

Although the model with Personalized-SDS, i.e, \textbf{P-SDS}, demonstrates significant improvements compared to standard \textbf{SDS}, the absence of geometric guidance from normals results in broken avatars. 
To this end, we use normal guidance in our \textbf{PNA-SDS} formulation Eq.~\ref{equ:pna_sds}, which further helps preserve the geometry as well as the key features in the original avatar. 
Yet, the random sampling of the timestep $t$ leads to a diminished resolution of fine details, underscoring the necessity for an annealing process. 

Additionally, we illustrate the ineffectiveness of the annealing strategy proposed by HiFA \cite{zhu2023hifa} (termed as \textbf{PNA-SDS + HiFA annealing}) when applied to our method. 
Here, we set a constant value for $k$ and deterministically select the value of diffusion timestep $t$ based on the iteration step. 
This is ineffective for our approach as once $t<k$, the personalized model does not contribute to the score, resulting in a complete breakdown of the avatar.

In striking contrast, our full method, termed \textbf{Ours}, delivers high-quality edits while preserving the crucial details of the pre-trained human avatar. 
The CLIP scores for the ablation study (Tab.~\ref{tab:clip_ablation}), further prove that our complete method outperforms all design alternatives in terms of alignment with the textual prompts and overall visual quality.

\input{tables/clip_ablation}

%% file: figures/fig_comparisons.tex
\begin{figure}[t]
    \centering
    \includegraphics[width=0.98\linewidth]{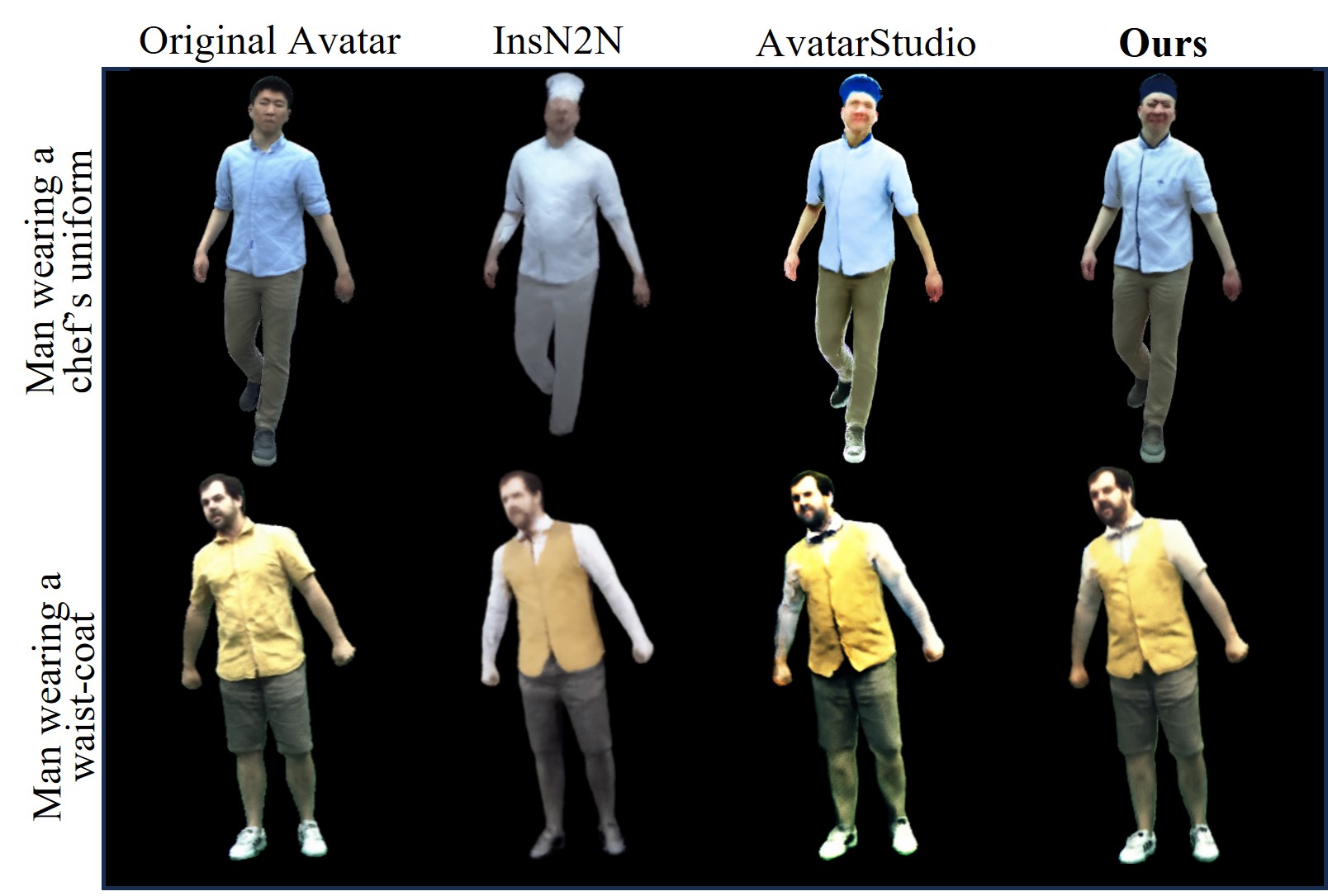}
    \vspace{-5px}
    \caption{
        \textbf{Qualitative Comparison.} 
         Our approach preserves the subject's characteristics and produces visually compelling edits that align with the prompts.
    }
    \label{fig:comparison}
\end{figure}

\begin{table}[t]
    \centering
    \footnotesize{
        \begin{tabular}{cccc}
            \toprule
            & InsN2N~\cite{instructnerf2023} & AvatarStudio~\cite{10.1145/3618368}  & \textbf{Ours} \\ \hline
            Q1  & 9.6 & 12.8& \textbf{77.6} \\ \midrule
            Q2  & 12& 24 & \textbf{64} \\ \midrule
            Q3 & 14.4 & 8  & \textbf{77.6} \\ \midrule
            Q4& 11.2  & 10.4 & \textbf{78.4} \\ \bottomrule
        \end{tabular}
    }
    \vspace{-5px}
    \caption{\textbf{User Study.} Results from our study with 25 participants. Our approach outperforms AvatarStudio and InstructNeRF2NeRF (InsN2N) by a large margin.}
    \label{tab:userstudy}
\end{table}
%

%% file: tables/quantitative.tex
\begin{table}[t]
    \footnotesize
    {
    \centering
    \begin{tabular}{cccc}
        \toprule
        \textbf{} & InsN2N~\cite{instructnerf2023} & AvatarStudio~\cite{10.1145/3618368} & \textbf{Ours} \\
        \midrule
        CLIP Similarity $\uparrow$  & 0.1656 & 0.1412& \textbf{0.2098} \\
        \midrule
        FID Score $\downarrow$  & 98.932 & 116.732 & \textbf{80.74}\\
        \bottomrule
    \end{tabular}
    }
    \caption{
    We adopt CLIP text-image Direction Similarity to assess the alignment of the edits with the text within the CLIP space. Additionally, we report the FID scores to evaluate the preservation of identity and geometric accuracy of the avatars.
    }
    \label{tab:quantitative}
\end{table}

%% file: tables/clip_ablation.tex
\begin{table}[t]
\footnotesize
{
\centering
\begin{tabular}{@{}cccccc@{}}
\toprule
SDS & NA-SDS & P-SDS & PNA-SDS & PNA+HiFA & \textbf{Ours} \\ \midrule
0.1153 & 0.2256 & 0.1844 & 0.2005 & 0.1159 & \textbf{0.2409} \\ \bottomrule
\end{tabular}
}
\caption{CLIP-Similarity scores corresponding to the ablations presented in Fig.5. Note that each of our design choices leads to an improvement compared to the baselines.}
\label{tab:clip_ablation}
\end{table}

%% file: sections/7_conclusion.tex
%
%

\section{Conclusion} \label{sec:conclusion}
In this work, we tackled the problem of intuitive editing of 3D neural avatars where a user can specify a desired edit via text prompts.
Our method then automatically adjusts the shape and appearance of the neural avatar to fulfill the user's demands while maintaining the subject's visual coherence. 
%
At the technical core, we propose a Personalized Normal-Aligned Score Distillation Sampling and a windowed timestep annealing to ensure space-time consistency in edits and high visual fidelity.
Our results demonstrate a clear step towards more intuitive, high-fidelity 3D neural avatar editing and outperform respective competing methods.
While TEDRA facilitates the intuitive editing of 3D neural human avatars with space-time consistency and superior visual quality, it faces challenges with long training times. Future work will focus on enabling more detailed edits while reducing resource demands. 

%

%% file: sections/X_suppl.tex
\clearpage
\setcounter{page}{1}
\maketitlesupplementary

%
%
\appendix
%
%
This supplemental document provides further information about the implementation details (Sec.~\ref{sec:suppl_implementation}). The additional results (Sec.~\ref{sec:supp_results}) showcase testing on various subjects, free-viewpoint rendering capabilities, animation transfer demonstrating pose adaptability to multiple avatars, further ablation studies, and qualitative comparisons, highlighting the robustness and versatility of our approach. Finally, we address the limitations of our study and suggest potential directions for future research (Sec.~\ref{sec:supp_limitations}).

\input{suppl_sections/1_implementation_details}

\input{suppl_sections/2_results}

%% file: suppl_sections/1_implementation_details.tex
%
%
\section{Implementation Details} \label{sec:suppl_implementation}
We first provide more details about the dataset (Sec.~\ref{sec:suppl_dataset}), followed by more details concerning fine-tuning the diffusion model (Sec.~\ref{sec:supp_finetuning}), editing the avatar (Sec.~\ref{sec:supp_editing}), and our annealing strategy (Sec.~\ref{sec:supp_annealing}).
%
%
\subsection{Dataset} \label{sec:suppl_dataset}
The dataset adopted to train the deformable avatar representation consists of two parts: the DynaCap~\cite{habermann2021} dataset and the newly recorded sequences.
The DynaCap dataset consists of $5$ subjects wearing different types of apparel, performing diversified everyday motions.
In this paper, we take one representative sequence from the DynaCap dataset for training the deformable avatar model.
Notably, we follow the protocols mentioned in the TriHuman~\cite{zhu2023trihuman} and train the deformable avatar using the training splits provided by the DynaCap dataset.
Apart from the DynaCap dataset, we captured 3 new sequences to demonstrate the effectiveness of our model.
The sequence features $3$ Subjects wearing everyday clothing and engaging in various activities, including running, jumping-jack, boxing, and dancing.
The sequences are recorded in a multi-view studio with $120$ $4K$ cameras at a frame rate of $25$ fps.
Inspired by the protocol proposed by DynaCap Dataset, we recorded separate training and testing sequences with 27,000 and 7,000 frames. 
Specifically, we hold out $4$ cameras from different viewing directions as testing camera views.
Additionally, we annotate all the captured frames with 3D skeletal poses (generated with markerless motion capture software~\cite{captury}), and foreground segmentation masks (produced by the state-of-the-art background matting method~\cite{BMSengupta20}).
We will make the data and the annotations, publicly available for research use upon acceptance.
%
%
\subsection{Fine-tuning Details} \label{sec:supp_finetuning}
We start by rendering images at 1fps and 50 views from the pre-trained avatar, the rendered images are then used to fine-tune the U-net ($\hat{\zeta}_{\phi}$) and the text-encoder ($\Gamma$) of the Latent Diffusion Model (LDM) \cite{rombach2021highresolution}. 
We follow the fine-tuning strategy proposed by DreamBooth \cite{ruiz2022dreambooth}:
\begin{equation}
    \mathbb{E}_{x_i, s_i, \zeta, t} \left[ w_t \|\hat{{\zeta}}_{\phi}(\textbf{z}_{t,i}, \textbf{s}) - \mathcal{E}(\textbf{x}_i) \|^2
_2 \right],
\end{equation}
where $\textbf{z}_{t,i} = \alpha_t \mathcal{E}(\textbf{x}_i) + \beta_t\boldsymbol{\epsilon}$, $ \boldsymbol{\epsilon} \sim \mathcal{N}(0, I)$, $\textbf{x}_i$ is the rendered image and $\alpha_t, \beta_t$ control the noise schedule. We use $w_t = {\sigma^2}{\sqrt{1 - \sigma^2}} $ as proposed in Fantasia3D \cite{Chen_2023_ICCV} for appearance modeling. Once trained, this model can now generate images given the prompt 'a photo of a sks man/woman' in random poses and viewpoints.

To achieve pose and view-point control of the generated images we employ a pre-trained ControlNet~\cite{zhang2023adding} which is conditioned on normal-maps. TriHuman is capable of generating images of surface normals which are computed by positional derivatives of the SDF field. Along with the computed normals and an empty string as input, the ControlNet now acts as an encoder to provide pose and view control over generated images of the fine-tuned LDM. 

Using this strategy, our fine-tuned model can also generalize avatar editing to novel views and poses. The fine-tuning is performed for 20,000 iterations with a batch size of 30, and the learning rate is set to 1e-6.
%
%
\input{figures_suppl/supp_ablation_1}
\subsection{Editing Details} \label{sec:supp_editing}
During the editing phase, both the latent diffusion models and ControlNets are frozen, and only the TriHuman model is optimized using the proposed score distillation termed PNA-SDS (Personalized Normal-Aligned Score Distillation Sampling) as defined in Eq.~\ref{equ:pna_sds}. 

Since we use pre-computed normals for both the pre-trained and personalized LDM, we set the ControlNet conditioning scale to 0.5 and 1.0 respectively. This allows the pre-trained LDM to facilitate geometrical changes toward the targeted edit. Fig.~\ref{fig:rebuttal_fig} (a) shows the impact of ControlNet conditioning scale on samples from pre-trained LDM.

The hyperparameters \( v = 0.3\) and \( k = 750\) are empirically chosen to optimize the balance between preserving the original identity of the avatar and achieving the desired edits. The impact of these values can be found in Section 4.3 of SINE \cite{Zhang2022SINESI}.

For a sequence of length 1k frames, we optimize the TriHuman model for 50k iterations with a learning rate of 1e-4 on an NVIDIA A100 GPU. We utilize classifier-free guidance with $w = 20$.

%
%
\subsection{Time-step Annealing Details} \label{sec:supp_annealing}
With reference to Eq.~\ref{time_step_annealing} we set the maximum and minimum diffusion timesteps as $t_{\mathrm{max}} = 980$ and $t_{\mathrm{min}} = 20$,  with a window size of $w = 500$. Initially, this configuration yields $t_1 = 980$, $t_2 = 480$, and a blending threshold of $k = 730$. The annealing process ceases once $t_1$ reaches 500 to prevent further increases in blurriness. Fig.~\ref{fig:annealing} shows a graphical representation of the proposed annealing strategy.
%
\input{figures/fig_annealing}
%
\par 
The windowed root annealing prioritizes larger timesteps \(t\) early in the training process, which, akin to diffusion models, establishes the target semantics quickly.
In Fig.~\ref{fig:rebuttal_fig} (b), the windowed root annealing method shows the formation of a 'cap' by prioritizing larger timesteps \(t\) early in training. As training progresses, \(t\) gradually decreases, refining fine details without losing them, a risk present at higher timesteps.

%% file: figures_suppl/supp_ablation_1.tex
\begin{figure}[H]
    \centering
    \vspace{-1em}
    \includegraphics[width=0.95\linewidth]{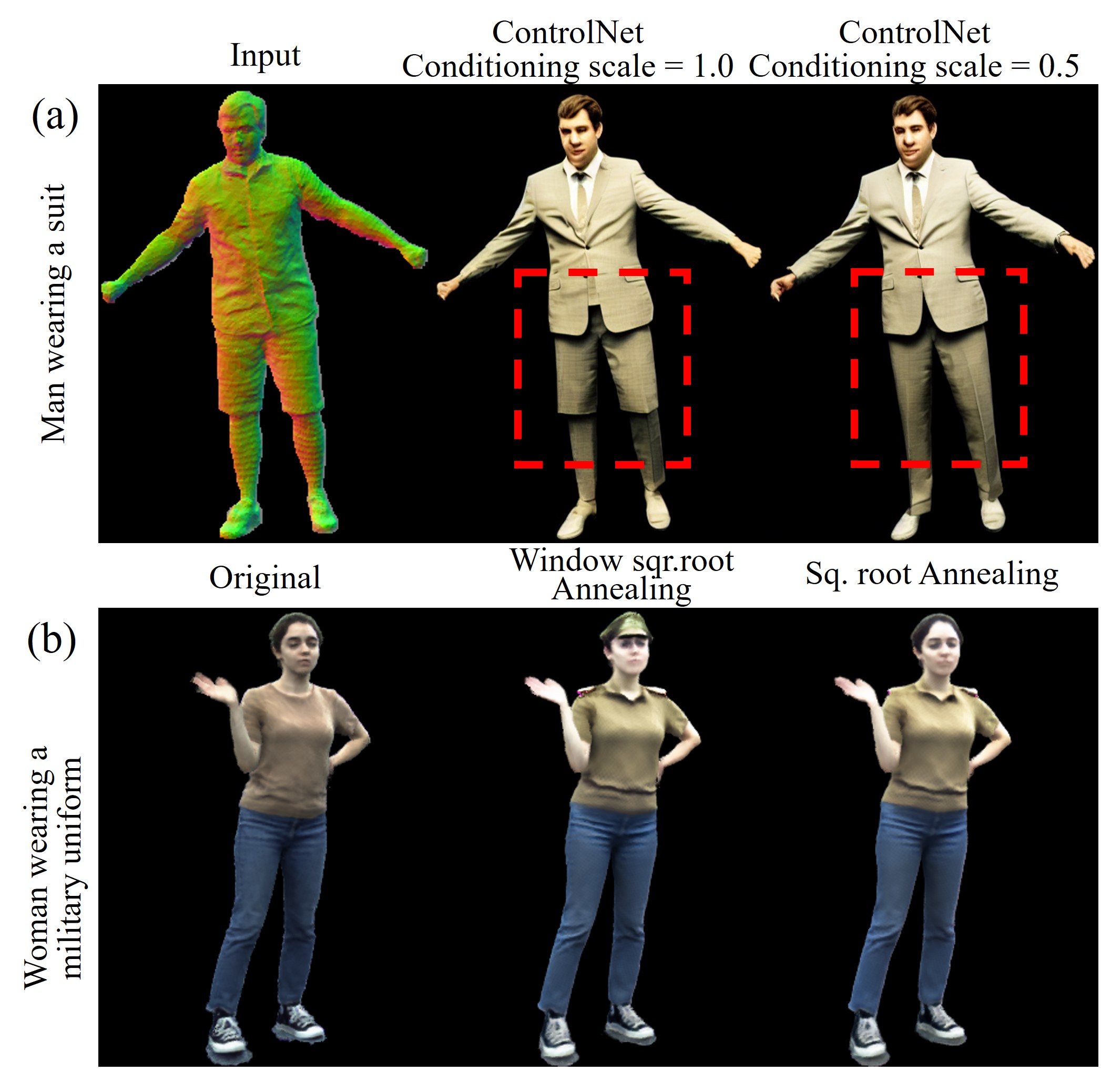}
    \caption{More ablative results on the  ControlNet conditioning scale, and the windowed root annealing method, please \textbf{zoom-in} to see the details.}
    \label{fig:rebuttal_fig}
\end{figure}

%% file: figures/fig_annealing.tex
%
%
\begin{figure}[h]
  \centering
  \includegraphics[width=0.99\linewidth]{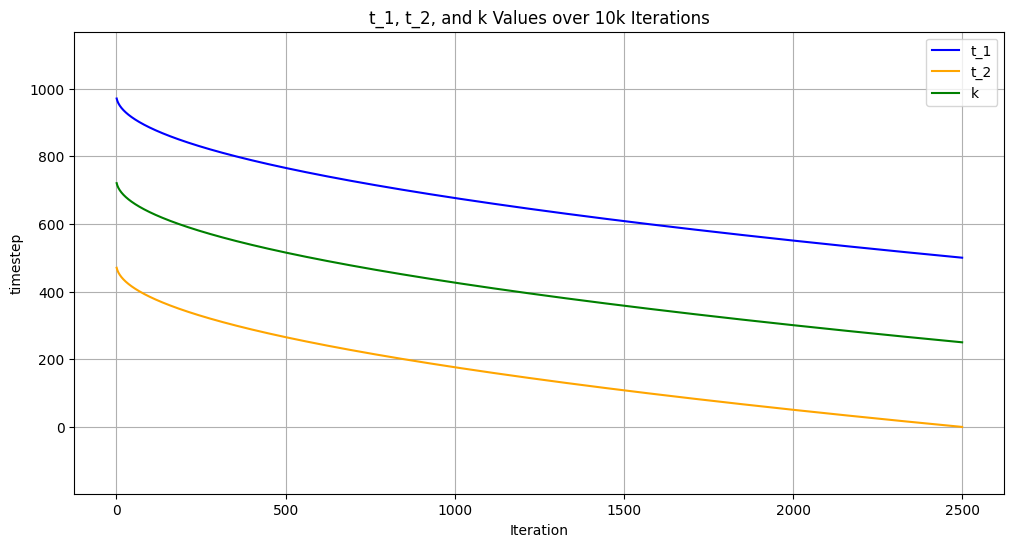}
  \caption
  {
    The figure shows the annealing of timesteps using the proposed window-root timestep annealing strategy for 10k iterations. The timestep \(t\) is randomly sampled within the shown window. As per Eq. 7 if \(t > k\) Then the scores from both pre-trained LDM and personalized LDM are used else only the scores from pre-trained LDM are used.
  }
  \label{fig:annealing}
\end{figure}
%

%% file: suppl_sections/2_results.tex
\section{Additional Results} 
\label{sec:supp_results}
The results section provides a comprehensive overview of the study's findings. 
It demonstrates the effectiveness of the methodology through enhanced editing results across a diverse range of subjects (Sec.~\ref{sec:supp_more_subjects}), underscoring its versatility. 
Additionally, the exploration of free-viewpoint rendering enriches visual representation (Sec.~\ref{sec:supp_free_view}), offering new perspectives on edited subjects, and the animation section (Sec.~\ref{sec:supp_animation}) showcases pose transfer adaptability to multiple avatars, highlighting the robustness of our approach. Further ablations reveal insights into variable impacts on the editing process (Sec.~\ref{sec:supp_ablations}).
Finally, we present additional comparisons of our method with state-of-the-art methods (see Sec.~\ref{sec:supp_comparisons}), highlighting advancements and limitations.

\subsection{More Subjects}
\label{sec:supp_more_subjects}
\input{figures_suppl/suppl_gallery_0}
\input{figures_suppl/suppl_gallery_1}
\input{figures_suppl/suppl_gallery_2}

Fig.~\ref{fig:suppl_results_id0}-\ref{fig:suppl_results_id2} provide more qualitative results on multiple subjects.
Our approach introduces a novel framework for generating visually pleasing edits guided by textual prompts across various contexts. 
The second row of Fig.~\ref{fig:suppl_results_id0} showcases how our system adeptly adjusts the appearance of gloves based on the provided textual guidance. 
Moreover, our method demonstrates a remarkable ability to target and modify specific regions as directed by the input text. 
For instance, the third row of Fig.~\ref{fig:suppl_results_id0} illustrates the versatility of our method, showcasing geometric alterations prompted by instructions such as "woman wearing a bicycle helmet". 
This demonstrates the proficiency of our system in interpreting complex textual instructions and creating visually appealing edits as a result.
Critical to our approach is the maintenance of subject consistency and coherence across both three-dimensional structure and temporal progression. 
This is substantiated by the consistency observed in our supplementary video evidence, reinforcing the reliability and efficacy of our method in generating visually consistent outcomes. 

In summary, our method excels in generating captivating visual edits driven by textual prompts, offering a flexible and intuitive approach to manipulating images across various scenarios.

\subsection{Free-viewpoint Rendering}
\label{sec:supp_free_view}
\input{figures_suppl/free_view_0}
\input{figures_suppl/free_view_1}
\input{figures_suppl/free_view_2}

Fig.~\ref{fig:suppl_results_free_id0}-\ref{fig:suppl_results_free_id3} presents the free-viewpoint renderings of the edited avatars. 
These results affirm that our approach maintains consistency across different viewpoints and time frames during the editing process.

\input{figures_suppl/animations}
\subsection{Animation}
\label{sec:supp_animation}
We demonstrate the ability to transfer poses from one character to multiple avatars, as shown in Fig.~\ref{fig:animations}. 
The results not only confirm the method's ability to perform versatile edits but also its adaptability to novel poses. This adaptability is particularly challenging within the domain of photorealistic 3D human avatars, highlighting the robustness of our approach.

\subsection{Additional Ablations}
\label{sec:supp_ablations}
We conduct qualitative ablations on one more identity with a different prompt as shown in Fig.~\ref{fig:suppl_ablations}. The terms used for different settings are as follows:

\textbf{SDS}: Score Distillation Sampling using only the pre-trained latent diffusion. 

\textbf{NA-SDS}: Normal Aligned SDS using pre-treined LDM and ControlNet.

\textbf{P-SDS}: Personalized SDS using fine-tuned/personalized and pre-trained LDMs.

\textbf{PNA-SDS}:Personalized Normal Aligned SDS using fine-tuned/personalized and pre-trained LDMs with pre-trained ControlNet conditioning (ours).

\textbf{PNA-SDS + HiFA Annealing}: PNA SDS with the diffusion timestep annealing strategy proposed by HiFA \cite{zhu2023hifa}.

\textbf{PNA-SDS + our Annealing}: PNA SDS along with our window root timestep annealing (our full method).

\input{figures_suppl/supp_comparisons}

The results clearly indicate the effectiveness of our comprehensive method, achieving high-quality edits while maintaining the essential details and dynamics of the pre-trained human avatar.
\input{figures_suppl/supp_ablation}
 
\subsection{Additional Qualitative Comparisons}
\label{sec:supp_comparisons}
In this section, we conducted more qualitative comparisons against competing approaches. 
Fig.~\ref{fig:suppl_comparisons} illustrates that our method maintains subject consistency while preserving clothing deformations. 
In contrast, 
Avatar Studio~\cite{10.1145/3618368} produces over-saturated and excessively smoothed results due to its limited subject information. 
Conversely, Instruct Nerf2Nerf~\cite{instructnerf2023} exhibits lower visual quality and reduced temporal consistency. Please refer to the supplementary video for more dynamic results.

\section{Limitations and Future Work}\label{sec:supp_limitations}
TEDRA significantly advances text-driven 3D avatar editing, providing compelling and coherent modifications. However, it struggles to recover fine facial details, like eyes, particularly because latent diffusion models struggle to sample full-body images with high-quality facial details. Our method's mask-based ray sampling restricts significant deviations in clothing from the pre-trained avatar model. Additionally, our method's dependency on per-prompt optimization and its intensive GPU requirements highlight areas for efficiency improvements.

Further, TEDRA needs data from a multi-view studio, limiting its accessibility. Exploring monocular setups for multi-view edits or developing dynamic, implicit representations of novel humans from text prompts offers promising directions for future research.

%% file: figures_suppl/suppl_gallery_0.tex
\begin{figure*}[t]
    \centering    \includegraphics[width=1.0\textwidth]{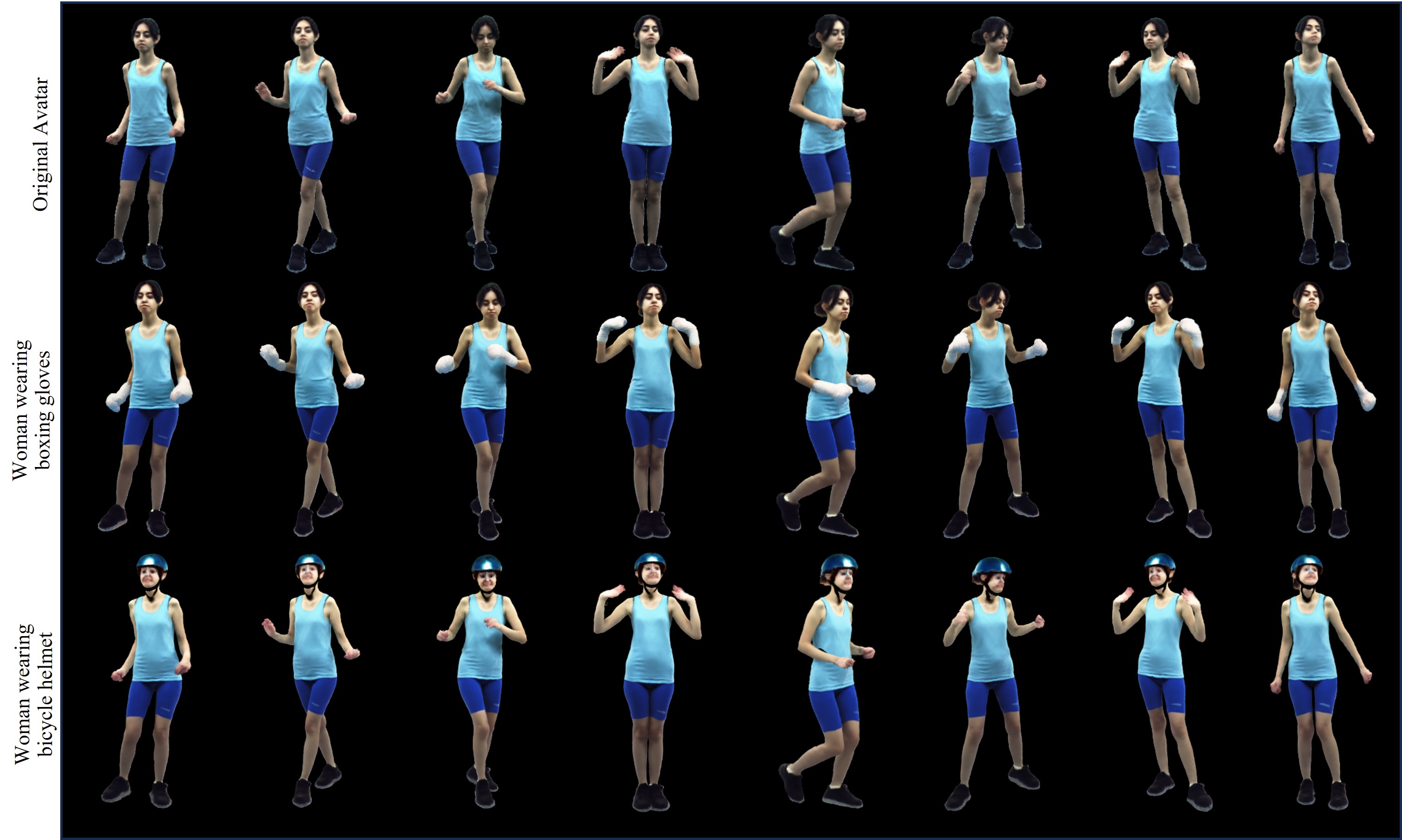}
    \caption{\textbf{Qualitative Results.} We present the text-based editing results. 
    We recommend the readers to \textbf{zoom in} to better view the details. 
    }
    \label{fig:suppl_results_id0}
\end{figure*}

%% file: figures_suppl/suppl_gallery_1.tex
\begin{figure*}[t]
    \centering    \includegraphics[width=0.99\textwidth]{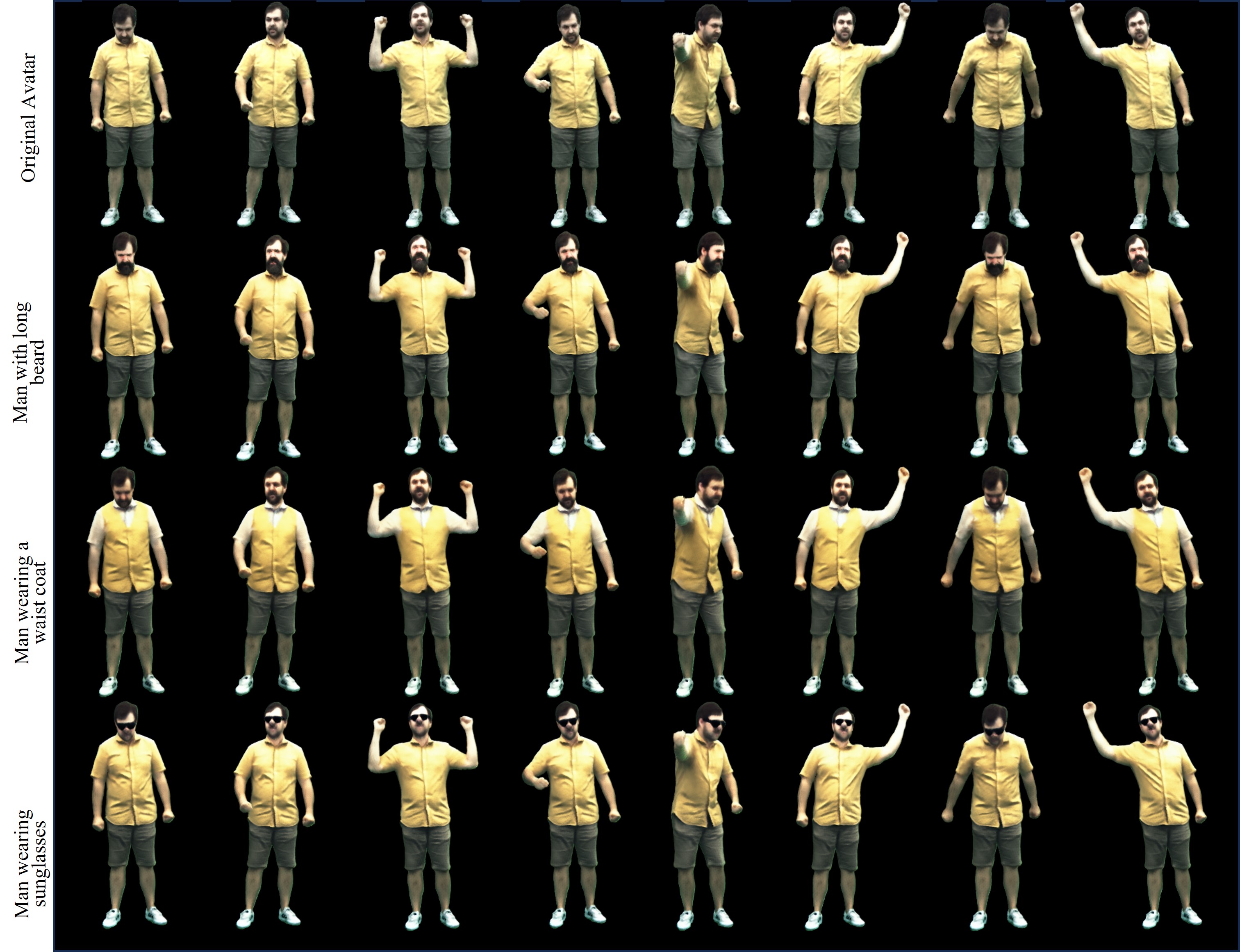}
    \caption{\textbf{Qualitative Results.} Our method demonstrates the capability to execute diversified contextual edits on photo-real avatars. These edits encompass various alterations such as adjusting length of the beard, as well as more localized changes that target specific.  
    We recommend the readers to \textbf{zoom in} for better viewing of the details.
    }
    \label{fig:suppl_results_id1}
\end{figure*}

%% file: figures_suppl/suppl_gallery_2.tex
\begin{figure*}
    \centering    \includegraphics[width=0.99\textwidth]{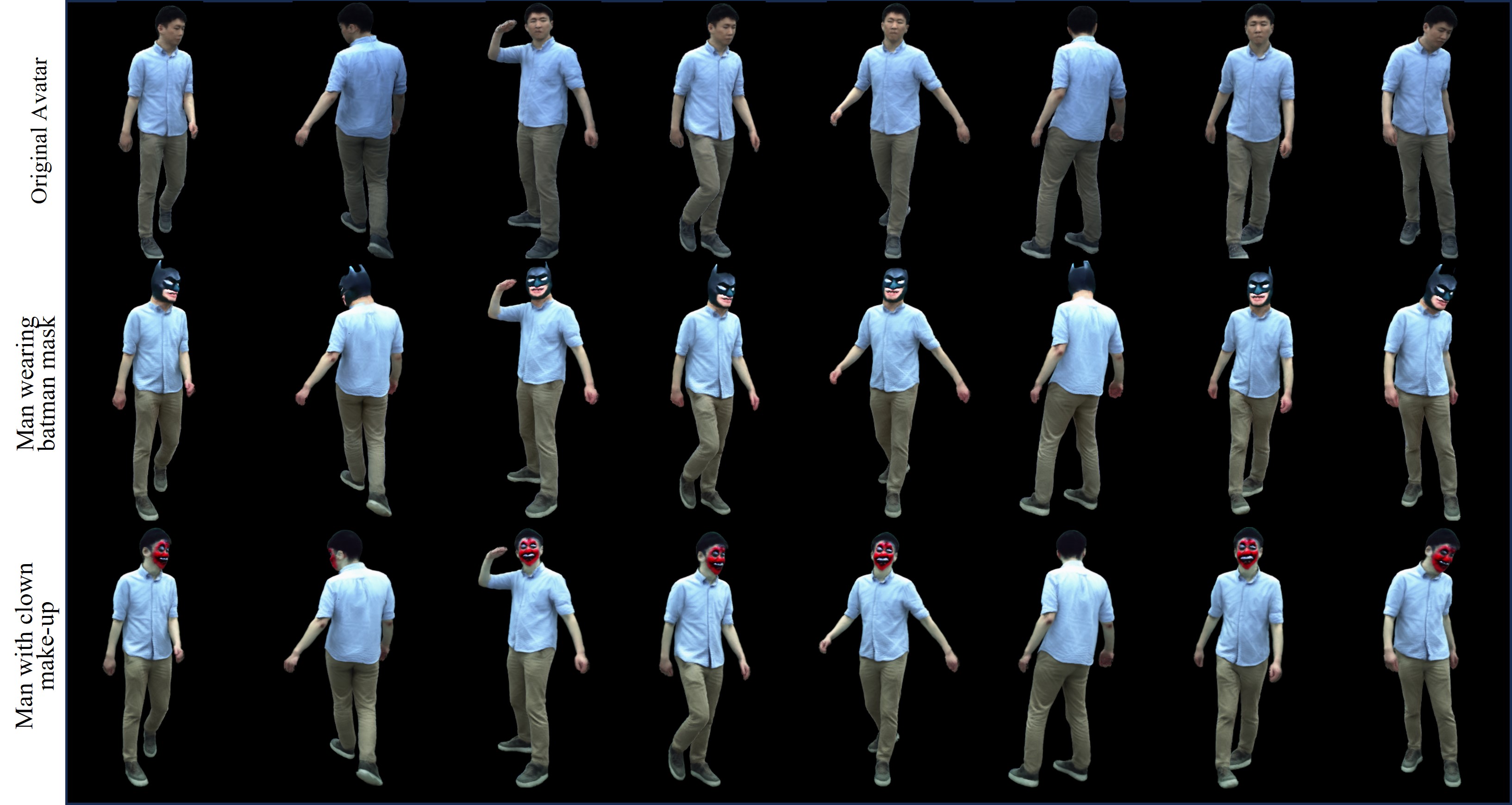}
    \caption{\textbf{Qualitative Results.}  
    Our approach generates captivating visual edits guided by textual prompts across various contexts. 
    We recommend the readers to \textbf{zoom in} for better viewing of the details.
    }
    \label{fig:suppl_results_id2}
\end{figure*}

%% file: figures_suppl/free_view_0.tex
\begin{figure*}[t]
    \centering    \includegraphics[width=0.95\textwidth]{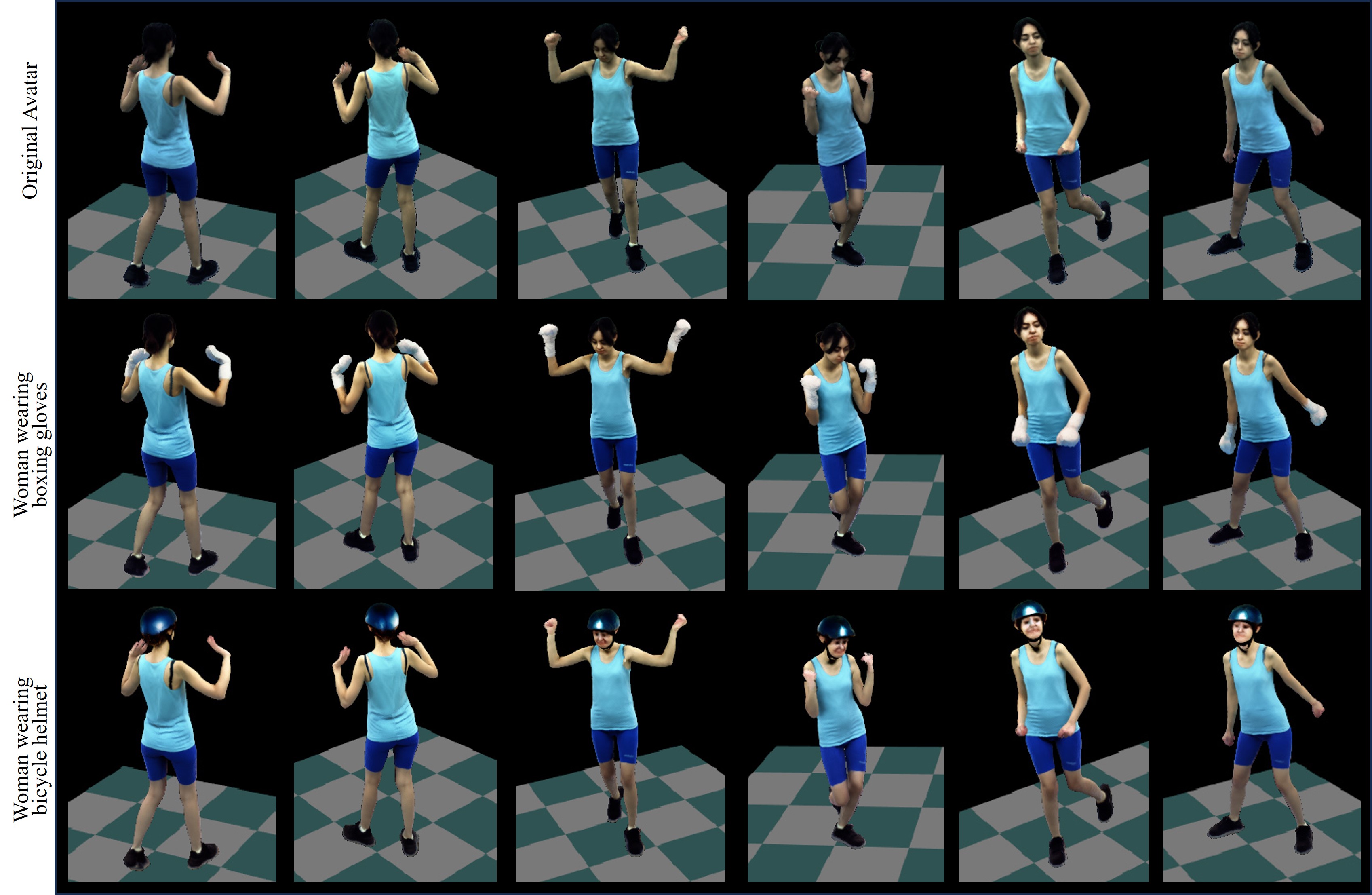}
    \caption{\textbf{Qualitative Results.} The free-viewpoint rendering results.
    We recommend the readers to \textbf{zoom in} to better view the details. 
    }
    \label{fig:suppl_results_free_id0}
\end{figure*}

%% file: figures_suppl/free_view_1.tex
\begin{figure*}[t]
    \centering    \includegraphics[width=0.90\textwidth]{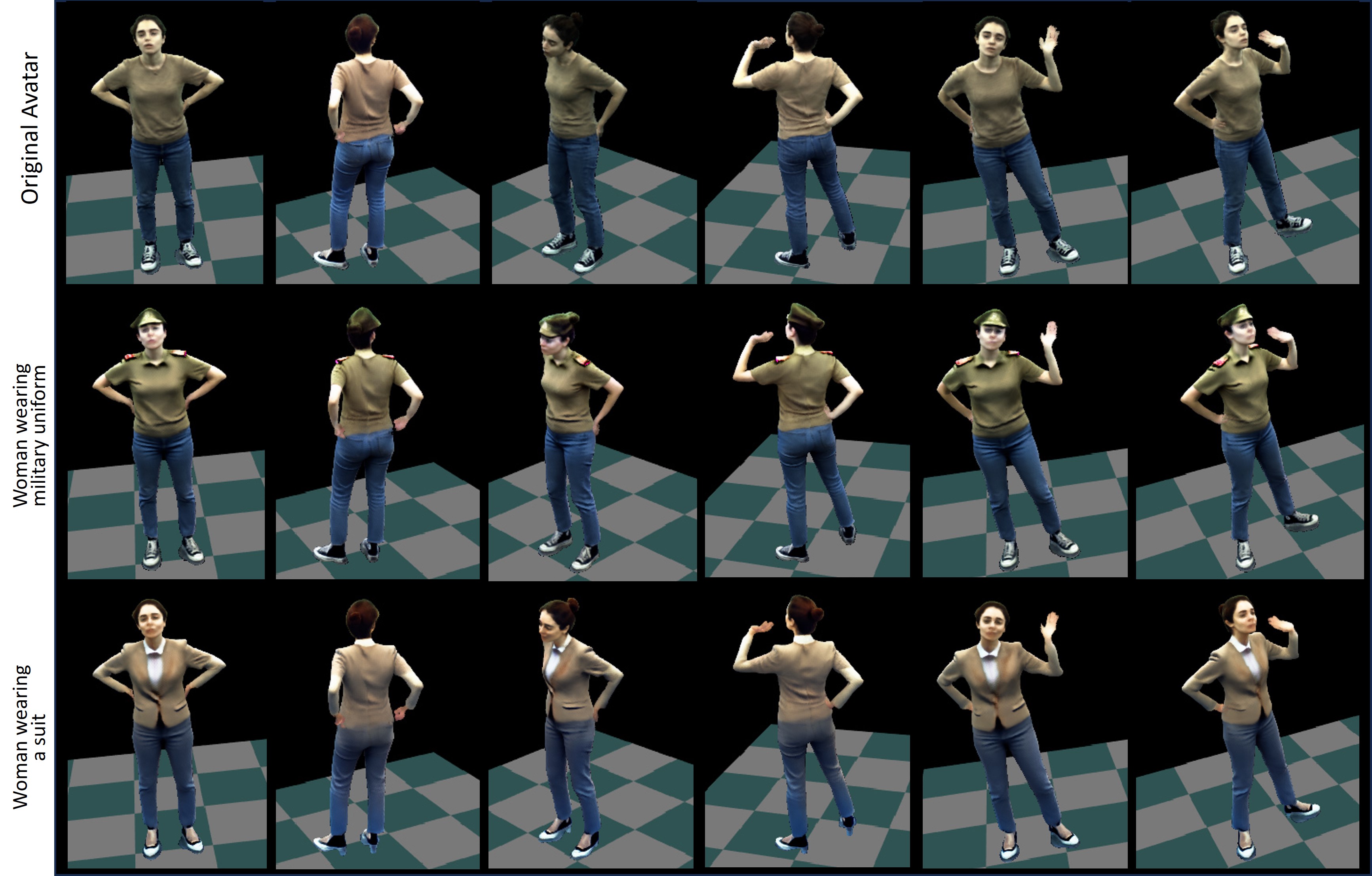}
    \caption{\textbf{Qualitative Results.} The free-viewpoint rendering results.
    We recommend the readers to \textbf{zoom in} to better view the details.
    }
    \label{fig:suppl_results_free_id1}
\end{figure*}

%% file: figures_suppl/free_view_2.tex
\begin{figure*}[t]
    \centering    \includegraphics[width=0.90\textwidth]{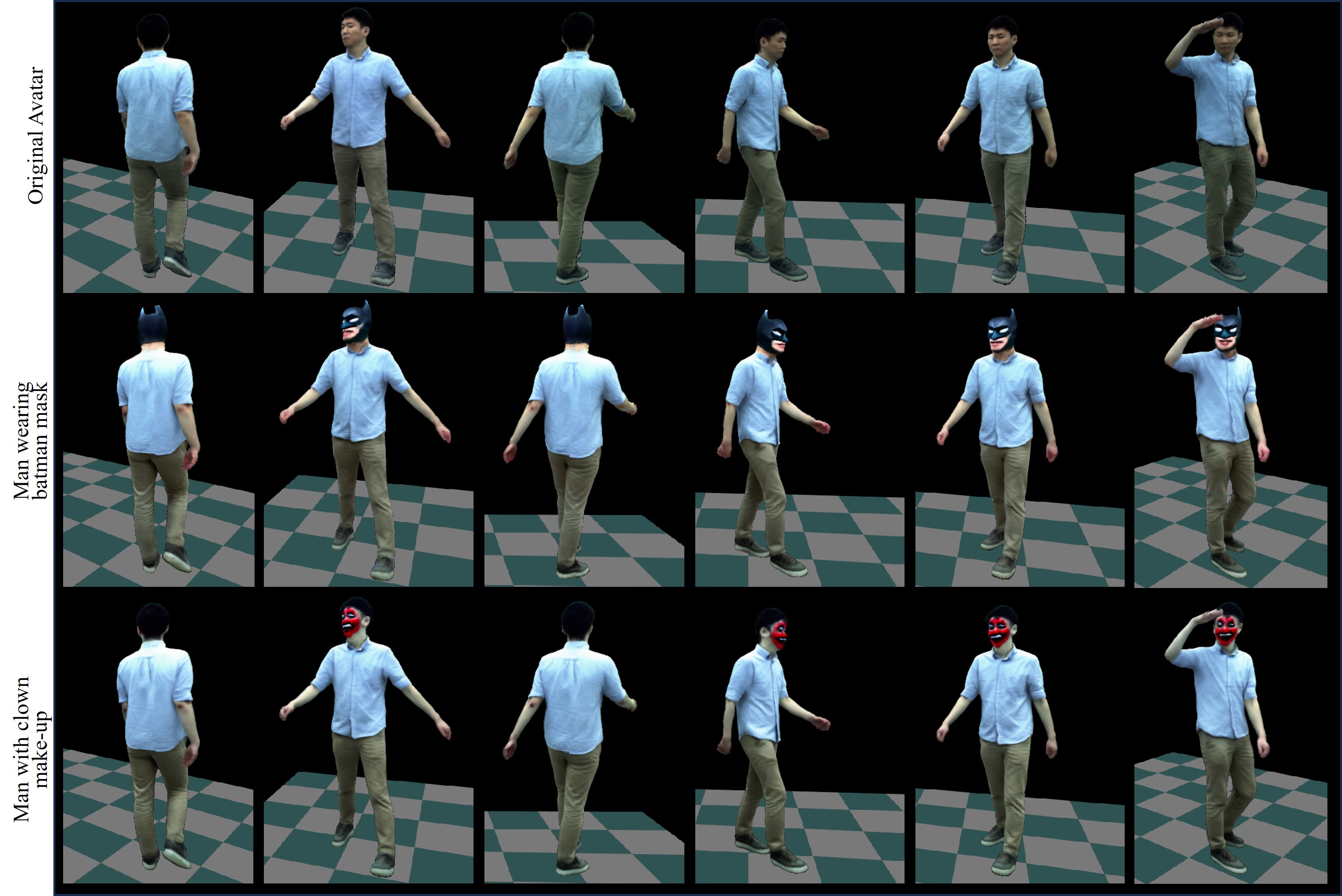}
    \caption{\textbf{Qualitative Results.} The free-viewpoint rendering results.
    We recommend the readers to \textbf{zoom in} to better view the details.
    }
    \label{fig:suppl_results_free_id3}
\end{figure*}

%% file: figures_suppl/animations.tex
\begin{figure*}[t]
    \centering    \includegraphics[width=1.0\textwidth]{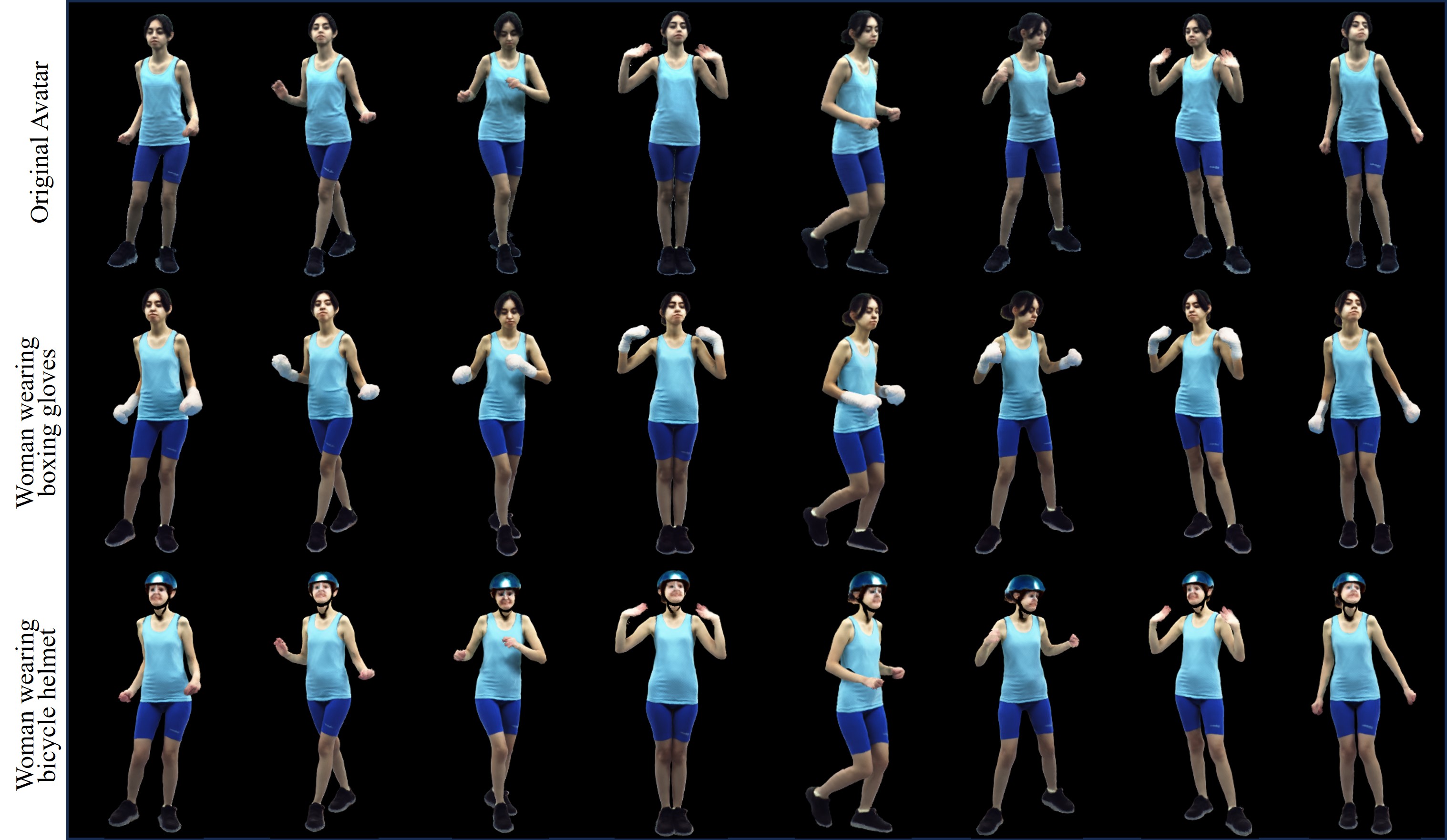}
    \caption{\textbf{Qualitative Results.} We present the results of avatar animation using novel poses.
    The first row shows the driving pose, followed by edited avatars driven by the same pose. 
    The results indicate that the edits are generalizable to novel poses.
    We recommend the readers to \textbf{zoom in} to better view the details. 
    }
    \label{fig:animations}
\end{figure*}

%% file: figures_suppl/supp_comparisons.tex
\begin{figure}[H]
    \centering    
    \includegraphics[width=0.95\textwidth]{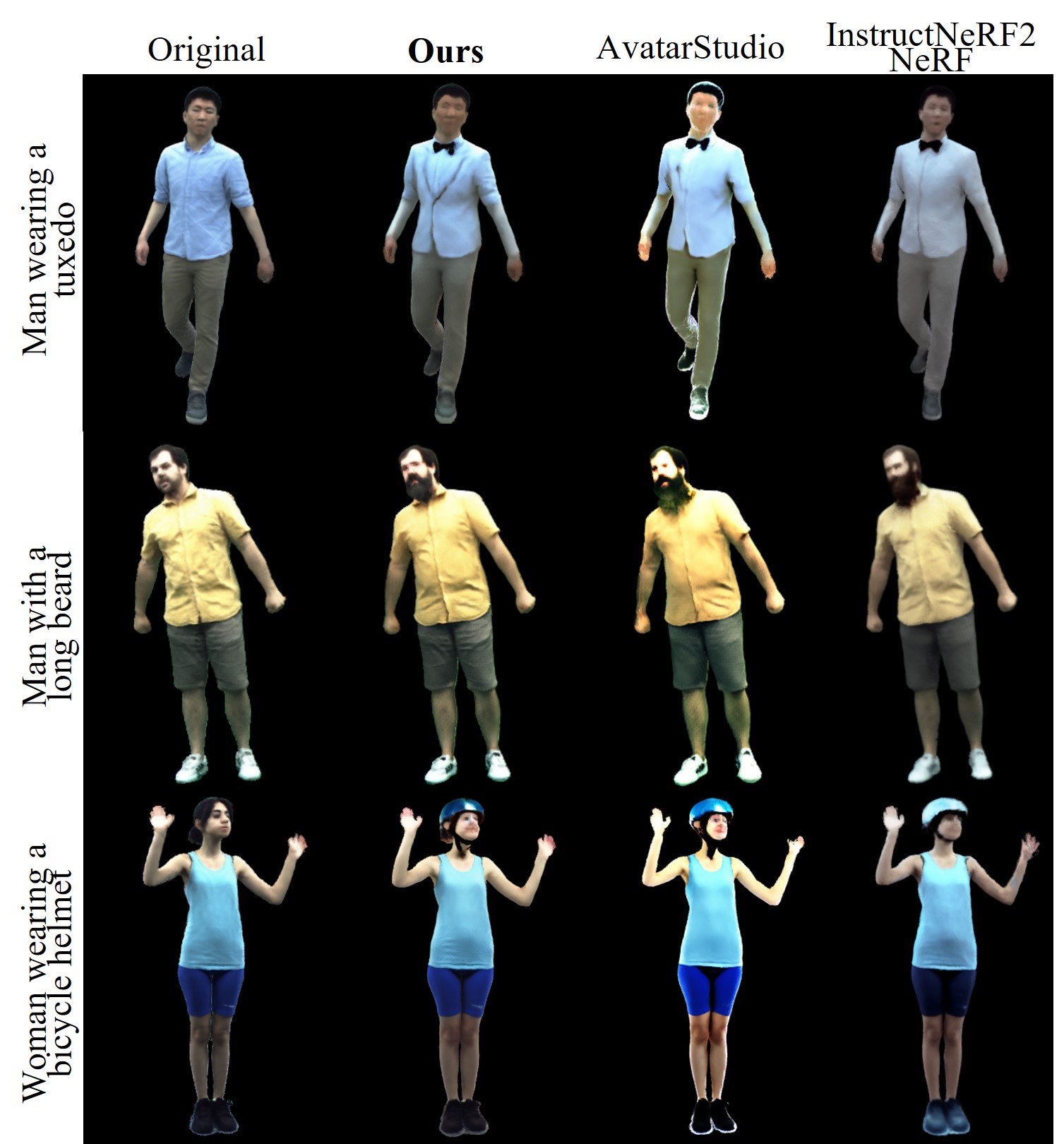}
    \caption{\textbf{Qualitative Comparisons.} We offer further comparisons involving AvatarStudio~\cite{10.1145/3618368} and InstructNeRF2NeRF~\cite{instructnerf2023}. Our findings indicate that the outcomes generated by alternative methods, specifically in columns 2 and 3, exhibit smoother surface details and lack consistency in maintaining subject coherence.
    %
    }
    \label{fig:suppl_comparisons}
\end{figure}


%% file: figures_suppl/supp_ablation.tex
\begin{figure*}[t]
    \centering    \includegraphics[width=1.0\textwidth]{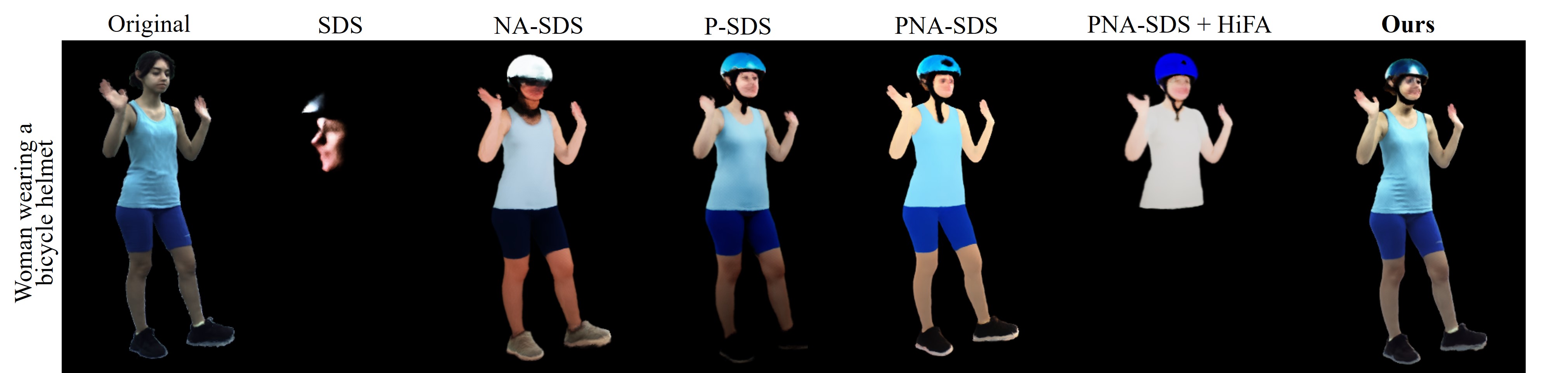}
    \caption{\textbf{Ablation Study.} We conduct a qualitative analysis, comparing our comprehensive approach to various design alternatives using the text prompt "a photo of a woman wearing a bicycle helmet". Our complete method successfully generates visually convincing modifications that maintain the crucial aspects of the original avatar. 
    %
    }
    \label{fig:suppl_ablations}
\end{figure*}